\PassOptionsToPackage{hyphens}{url}

\documentclass[9pt, twocolumn]{article}

\AtBeginDocument{%
  }

\usepackage{xcolor}

\usepackage{tikz}
\usetikzlibrary{positioning, arrows.meta}

\usepackage{array}
\usepackage{booktabs}
\usepackage{tabularx}
\usepackage{chngcntr}
\usepackage{placeins}
\usepackage{float}

\usepackage[utf8]{inputenc}
\usepackage{graphicx}
\usepackage{caption}  %
\usepackage{subcaption}
\usepackage{epigraph}
\usepackage[colorlinks=true, linkcolor=blue]{hyperref}
\usepackage{placeins}
\usepackage{wrapfig}
\usepackage{algorithm}
\usepackage[noend]{algpseudocode}
\usepackage{authblk}
\usepackage{times}
\usepackage{authblk}
\usepackage{amsfonts}
\usepackage{amsmath}
\usepackage{cleveref}
\usepackage{balance}
\usepackage{flushend}

\usepackage{algorithm}

\usepackage[top=2cm, bottom=2cm, left=2cm, right=2cm]{geometry}

\newcommand\blfootnote[1]{%
  \begingroup
  \renewcommand\thefootnote{}\footnote{#1}%
  \addtocounter{footnote}{-1}%
  \endgroup
}

\begin{document}

\title{Story2Board: A Training-Free Approach for Expressive Storyboard Generation}

\author[1]{David Dinkevich}
\author[1]{Matan Levy}
\author[1]{Omri Avrahami}
\author[2,3]{Dvir Samuel}
\author[1]{Dani Lischinski}
\affil[1]{Hebrew University of Jerusalem, Israel}
\affil[2]{OriginAI, Israel}
\affil[3]{Bar-Ilan University, Israel}

\date{}

\twocolumn[{
\maketitle

\vspace{-1.5em}
\centering
\includegraphics[width=\textwidth]{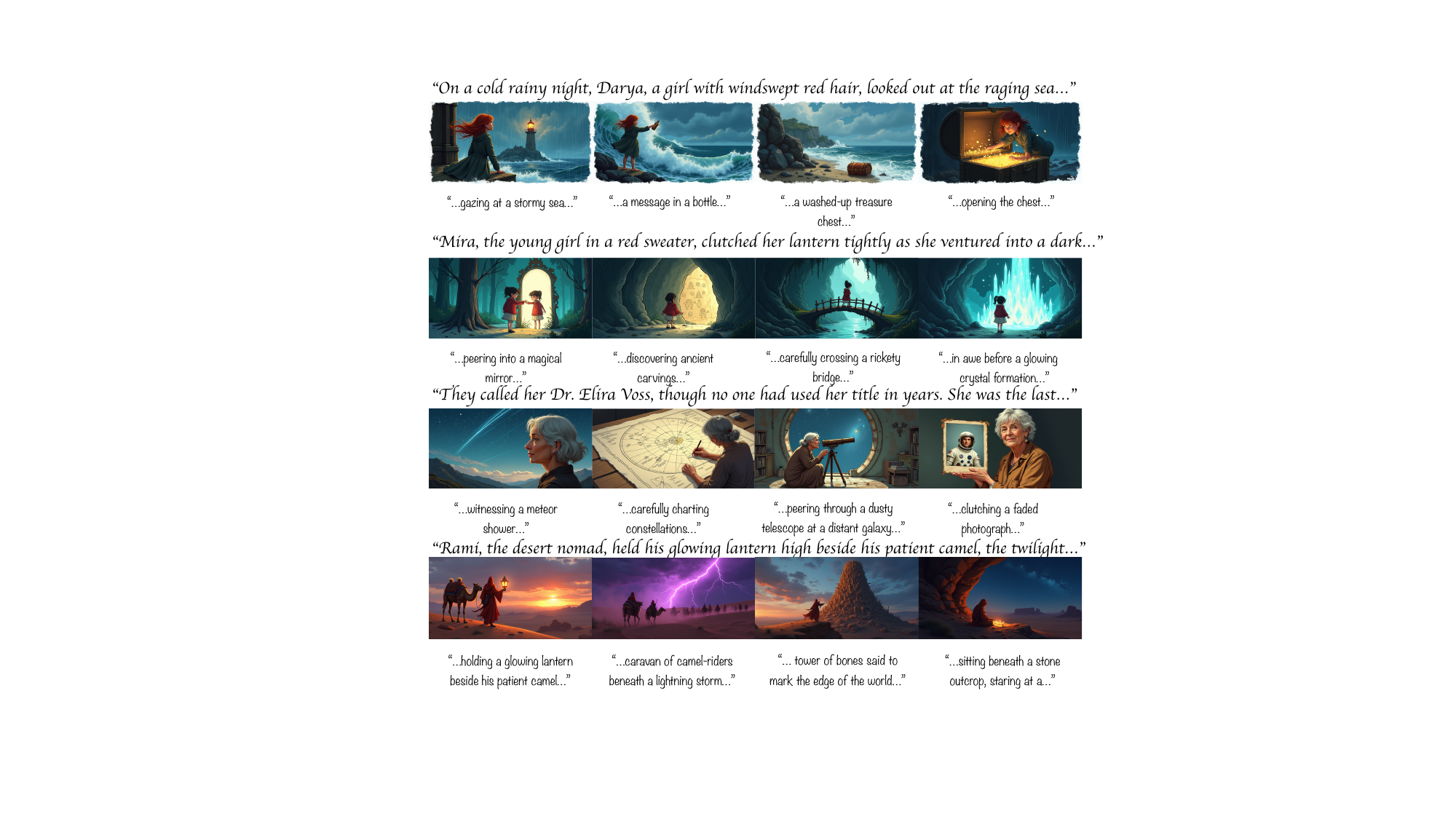}
\vspace{0.5em}
\captionof{figure}{
    \label{fig:teaser}
    \textbf{Story2Board} generates coherent multi-panel storyboards from a natural language prompt, maintaining subject identity while allowing dynamic changes in character pose, size, and position. Unlike prior work, it introduces a lightweight consistency mechanism that preserves the model’s generative prior, supporting rich, expressive storytelling without fine-tuning or architectural changes. Full story texts are available in the appendix (Section~\ref{sec:appendix-full-story-texts}).
}
\vspace{1em}
}]

\vspace{0.5em}
\small

We present \textit{Story2Board}, a training-free framework for expressive storyboard generation from natural language. Existing methods narrowly focus on subject identity, overlooking key aspects of visual storytelling such as spatial composition, background evolution, and narrative pacing. To address this, we introduce a lightweight consistency framework composed of two components: \textit{Latent Panel Anchoring}, which preserves a shared character reference across panels, and \textit{Reciprocal Attention Value Mixing}, which softly blends visual features between token pairs with strong reciprocal attention. Together, these mechanisms enhance coherence without architectural changes or fine-tuning, enabling state-of-the-art diffusion models to generate visually diverse yet consistent storyboards. To structure generation, we use an off-the-shelf language model to convert free-form stories into grounded panel-level prompts. To evaluate, we propose the \emph{Rich Storyboard Benchmark}, a suite of open-domain narratives designed to assess layout diversity and background-grounded storytelling, in addition to consistency. We also introduce a new \textit{Scene Diversity} metric that quantifies spatial and pose variation across storyboards. Our qualitative and quantitative results, as well as a user study, show that Story2Board produces more dynamic, coherent, and narratively engaging storyboards than existing baselines.

\section{Introduction}
\label{sec:intro}

Text-to-image (T2I) diffusion models ~\cite{ho2020denoising, ramesh2022hierarchical, Rombach2021HighResolutionIS, Saharia2022PhotorealisticTD, Podell2023SDXLIL} have rapidly transformed visual content creation, producing photorealistic and coherent images from natural language prompts with increasing reliability. Thanks to advances in open-source architectures and accelerated inference~\cite{Rombach2021HighResolutionIS, Saharia2022PhotorealisticTD}, these models have moved beyond research labs into creative workflows--illustrating children’s books, powering social media campaigns, and supporting early-stage animation pipelines~\cite{liu2023video, Zhang2023MagicBrush}. As these models become more accessible, they are increasingly adopted not just as tools for static image generation, but also as engines for visual storytelling~\cite{yang2024seed, he2025anystory}.

\blfootnote{Project page: \href{https://daviddinkevich.github.io/Story2Board}{https://daviddinkevich.github.io/Story2Board}}

Storyboards represent a natural next step in visual storytelling. More than just sequences of snapshots, they are structured visual narratives--compositions that evolve across time, depicting characters, environments, and emotional beats in a spatially and semantically coherent manner. Effective visual storytelling relies not only on visual fidelity, but also on principles of cinematic composition: scale, perspective, framing, and environmental grounding~\cite{block_visualstory}, as exemplified in Figure~\ref{fig:teaser}. Scenes such as a nomad dwarfed by a mount of bones, an empty beach under a stormy sky, or a girl illuminated by the glow of a treasure chest communicate narrative meaning through spatial arrangement and atmosphere--not just subject appearance. Capturing this expressive diversity requires T2I models to move beyond static character rendering and embrace dynamic scene construction. This includes varying viewpoint and depth, emphasizing background storytelling, and adapting character presentation to reflect the evolving arc of the narrative~\cite{filmmakersacademy_negspace, animatorisland_breathingroom}.

\begin{figure}[t]
\centering
\includegraphics[width=0.9\linewidth]{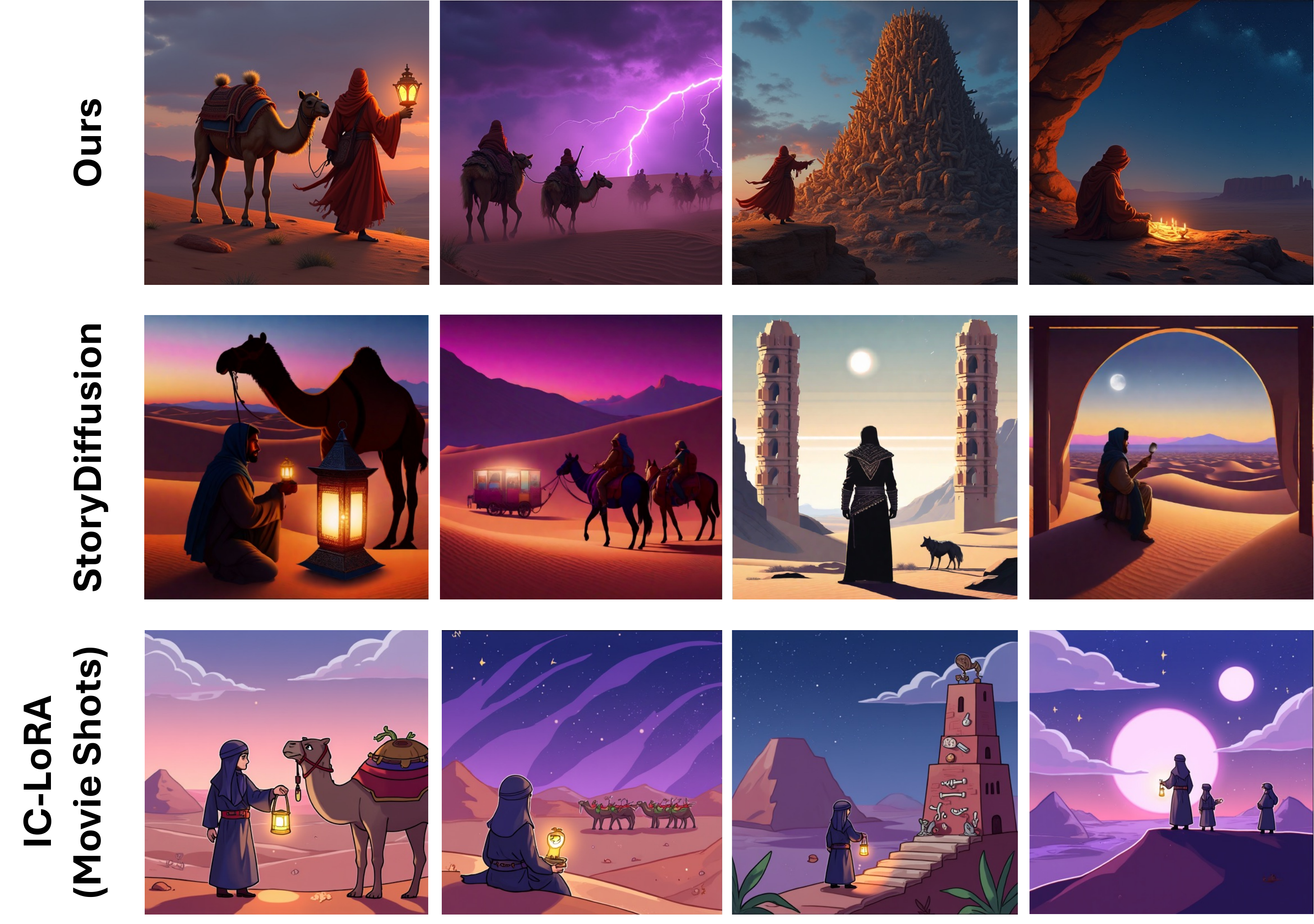}
\caption{
Comparative storyboard outputs from our method and two leading baselines, using the same input narrative. While baseline methods tend to center the character in every frame with limited variation in framing or environment, our method leverages cinematic principles--such as exaggerated scale, dynamic perspective, and environmental context--to convey narrative progression more expressively. Note, for instance, how the small scale of the character in the third panel of the top row enhances the sense of vastness of the tower of bones, reinforcing the emotional arc of the story.
}
\label{fig:motivation}
\end{figure}

Despite growing interest in automatic storyboard generation, current methods remain limited in their ability to produce visually compelling and narratively coherent image sequences. Several approaches focus narrowly on preserving character identity across frames--whether via reference-guided generation~\cite{tan2024ominicontrol, ye2023ip-adapter, Wei2023ELITEEV}, diffusion-based consistency~\cite{tewel2024consistory, he2024dreamstory, zhou2024StoryDiffusion}, or autoregressive modeling~\cite{liu2024storygen}--but often do so at the expense of compositional diversity. As illustrated in Figure~\ref{fig:motivation}, generated characters are typically centered, scenes may lack spatial depth, and prompts tend to follow rigid templates such as ``a photo of [character] in [setting].'' As a result, these storyboards resemble slideshows rather than expressive visual narratives.

To address these limitations, we propose a novel, training-free consistency framework that combines \textbf{Latent Panel Anchoring} and \textbf{Reciprocal Attention Value Mixing} to guide modern T2I models toward generating coherent and expressive storyboards. Crucially, our method does not constrain the model’s inherent generative capacity. Instead, it amplifies the in-context strengths of diffusion transformer (DiT) architectures by preserving a shared reference during denoising and softly blending appearance features between semantically aligned token pairs. This reinforces character identity and inter-panel coherence, while preserving the full compositional flexibility and visual richness of the base model. Importantly, our approach introduces no architectural changes or fine-tuning--offering token-level guidance that unlocks consistency without sacrificing diversity.

To interface with user input, we include a lightweight prompt decomposition step that converts natural-language stories into scene-level prompts using an off-the-shelf language model. This helps bridge freeform storytelling and visual generation, without requiring prompt engineering. The resulting method is compatible with state-of-the-art DiT-based models such as Stable Diffusion 3 and Flux~\cite{stablediffusion3, flux}, and examples of our outputs are shown in Figure~\ref{fig:teaser}.

While prior work has largely focused on character consistency and prompt alignment, little attention has been paid to evaluating a model’s ability to convey story through composition, atmosphere, and scene dynamics. Existing benchmarks~\cite{zhou2024StoryDiffusion, he2024dreamstory} are typically composed of short, templated prompts with minimal environmental detail and limited narrative variation. They do not assess whether a model can depict a character seated on a mossy log, silhouetted beneath a starry sky, or dwarfed by a looming structure. Nor do they challenge the model to vary a character’s size, pose, or placement across panels, or to omit the character entirely when appropriate. To fill this gap, we introduce the \textbf{Rich Storyboard Benchmark}, a curated collection of open-ended stories designed to probe layout flexibility, background storytelling, and expressive visual composition across a range of narrative settings.

To complement the benchmark, we propose a new \textbf{Scene Diversity} metric, which quantifies variation in a character’s appearance across a storyboard sequence. Specifically, it captures changes in scale, pose, position, and visibility, reflecting how fluidly a model adapts character presentation to serve evolving narrative demands. Unlike identity-focused metrics that reward visual repetition, Scene Diversity encourages expressive variation while maintaining recognizability, aligning more closely with the goals of cinematic storytelling. By explicitly measuring how a character is framed, integrated into the scene, or de-emphasized when appropriate, this metric offers a novel lens on visual narrative flexibility.

In summary, our contributions are threefold:
\begin{enumerate}
  \item We introduce a novel training-free consistency framework that combines Latent Panel Anchoring and Reciprocal Attention Value Mixing to enhance in-context coherence in diffusion transformer models. This enables expressive storyboards with consistent characters, dynamic layouts, and rich environmental composition--without compromising diversity or requiring model fine-tuning.
  \item We present the Rich Storyboard Benchmark, a suite of open-ended, visually grounded stories designed to evaluate layout flexibility, background detail, and narrative expressivity--dimensions underexplored in existing datasets.
  \item We propose a new metric, \textbf{Scene Diversity}, which quantifies variation in character pose, scale, and framing across panels, offering a more nuanced assessment of visual storytelling beyond identity preservation.
\end{enumerate}

\section{Related Work}
\label{sec:related}

Text-to-image (T2I) diffusion models ~\cite{ho2020denoising, ramesh2022hierarchical, Rombach2021HighResolutionIS, Saharia2022PhotorealisticTD, Podell2023SDXLIL} have revolutionized visual content generation, enabling high-quality synthesis from natural language prompts. Prominent recent models such as Flux~\cite{flux} and Stable Diffusion 3~\cite{Esser2024ScalingRF} exemplify the capabilities of large-scale transformer-based~\cite{vaswani2017attention} architectures in generating expressive, semantically grounded imagery. These models serve as a foundation for numerous methods for both consistent character synthesis and storyboard generation, two related but fundamentally distinct problem spaces.

Storyboard generation aims to produce sequences of images that together convey a narrative arc. The focus here is not solely on maintaining character identity, but on supporting dynamic compositions, evolving background elements, and expressive visual storytelling. In this space, StoryDiffusion \cite{zhou2024StoryDiffusion} introduces a consistency-aware attention module and a semantic motion predictor to guide narrative flow across frames. StoryGen \cite{liu2024storygen} introduces a learning-based autoregressive image generation model equipped with a vision-language context module, enabling coherent storyboard synthesis from freeform narrative input. DreamStory \cite{he2024dreamstory} similarly leverages a language model for prompt decomposition and employs a multi-subject diffusion architecture to preserve inter-character relationships across scenes. Other related efforts, such as IC-LoRA \cite{Huang2024InContextLF}, explore lightweight adaptation techniques to improve generation coherence across time steps, while OminiControl \cite{tan2024ominicontrol} introduces image-based conditioning to guide spatial layout and stylistic coherence throughout a narrative.

In contrast, consistent character generation~\cite{avrahami2024chosen, Tewel2024TrainingFreeCT} focuses on preserving the visual identity of a specific subject across multiple images. In this task, the character is typically the visual and semantic anchor of the composition, with background or narrative context playing a secondary role. Recent methods such as The Chosen One \cite{avrahami2024chosen}, ConsiStory \cite{Tewel2024TrainingFreeCT}, and IP-Adapter \cite{ye2023ip-adapter} manipulate internal representations--either through iterative prompt-based refinement, cross-image feature sharing, or external adapters--to maintain consistency across scenes. While some of these methods include the term “story” in their titles or describe sequential results, their primary concern remains identity fidelity. For example, in ConsiStory \cite{Tewel2024TrainingFreeCT}, consistency is measured almost exclusively through identity features, with little emphasis on narrative variation, layout dynamics, or background richness. This distinction is critical: consistent character generation centers the image around the character, whereas storyboard generation requires a broader representational range, where characters may appear small, partially occluded, shared with side actors, or absent altogether.

Our work targets storyboard generation, but diverges from existing approaches in three key ways. First, unlike methods that rely on training or model-specific finetuning, our pipeline is entirely training-free, and can be applied directly to pre-trained transformer-based diffusion models such as Flux and Stable Diffusion 3~\cite{Esser2024ScalingRF}. Second, our method is character-agnostic: we do not require character masks (e.g., via SAM~\cite{kirillov2023segment}) or reference tokens to locate or track characters across panels. Instead, we operate directly on token-level features, enabling a more flexible and general mechanism for maintaining consistency. Third, while some recent approaches rely on extended attention mechanisms to enable cross-image token sharing, we instead work within the original transformer architecture--intervening only at the level of value vector mixing between attended tokens. This allows us to retain the model’s compositional expressiveness and architectural simplicity, while still achieving strong identity and layout coherence across storyboard panels.

\newcommand{\methodoverviewcaption}{
Overview of our training-free storyboard generation pipeline. Given a natural language narrative (e.g., \textit{``Once upon a time, a boy set off on an adventure...''}), our method proceeds in three stages:
(1) An LLM-based ``Director'' decomposes the story into a shared \textit{reference panel prompt} and a sequence of \textit{scene-level prompts};
(2) A batch of $n$ two-panel images is generated, with the top half of each image conditioned on the (same) reference prompt and the bottom half on one of the scene prompts. During denoising, we apply \textit{Latent Panel Anchoring (LPA)}: after each transformer block, the latent representations $[R, p_i]$ evolve to $[R'_i, p'_i]$, and the top half of each latent is replaced with the version from the first batch element, denoted $R' = R'_1$, to ensure a synchronized anchor across scenes. Inside each transformer block, we also apply \textit{Reciprocal Attention Value Mixing (RAVM)} following the self-attention computation.
(3) The final denoised latents are decoded into two-panel images and cropped to retain only the bottom sub-panels as the final storyboard.
}

\begin{figure*}[t]
    \centering
    \includegraphics[width=1.0\textwidth]{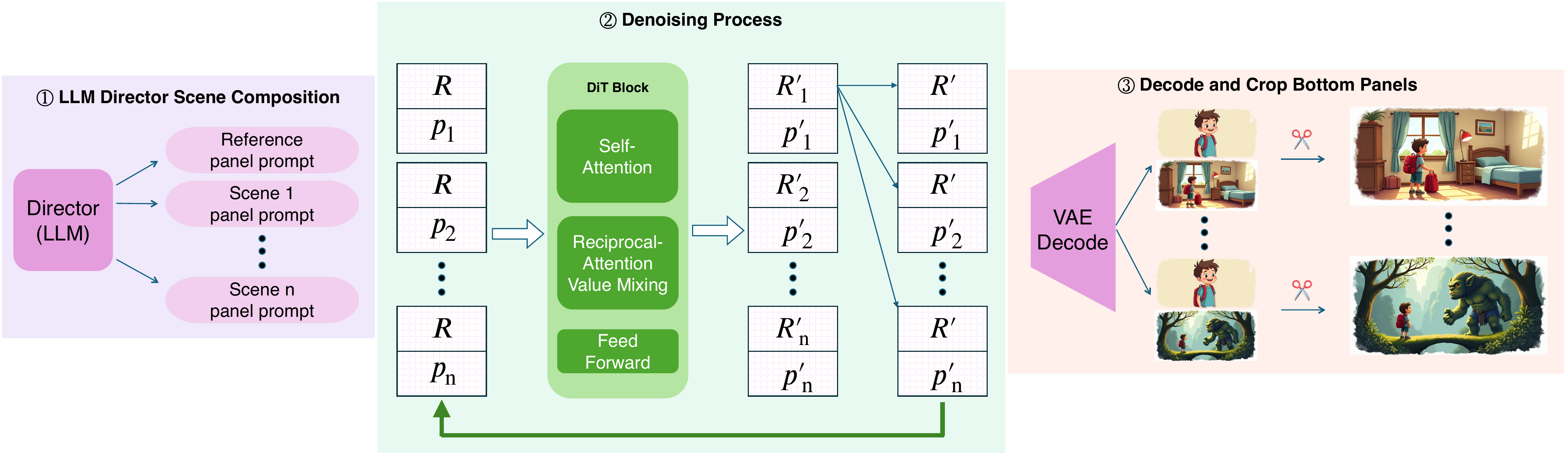}
    \caption{\methodoverviewcaption}
    \label{fig:method_overview}
\end{figure*}

\begin{figure}[t]
    \centering
    \includegraphics[width=0.8\linewidth]{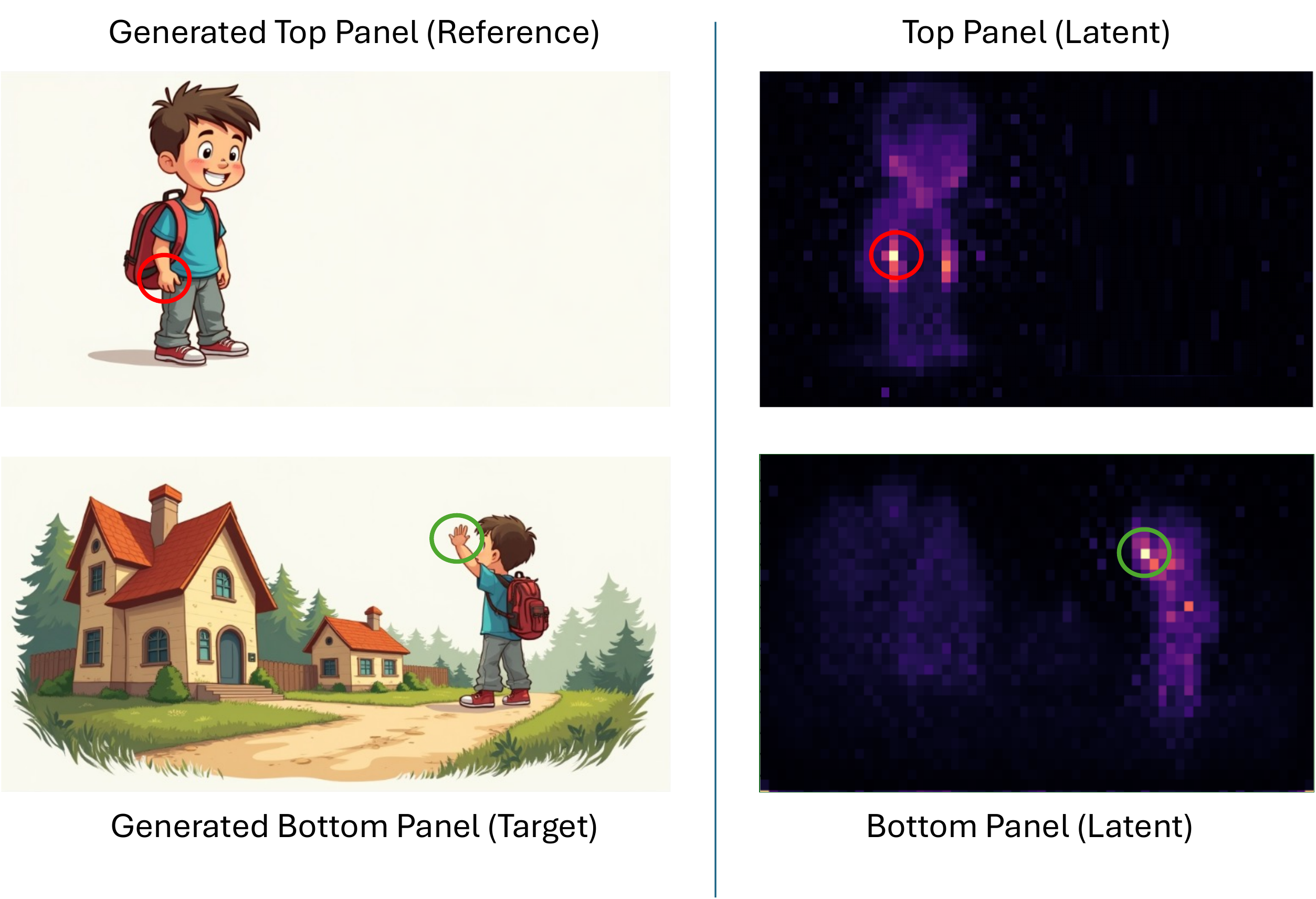}
    \caption{
    Visualization of \textit{Reciprocal Attention Value Mixing (RAVM)} in action. \textbf{Left:} A generated 2-panel output from our method, with the top panel serving as the shared reference. The red and green circles mark semantically corresponding character features (the hand) in the reference and target panels, respectively. \textbf{Right:} Heatmaps showing reciprocal attention scores at denoising step 12 of 28. Top-right: for each token in the top panel, we compute its reciprocal attention with the green-circled token in the bottom panel. Bottom-right: the reverse--each token in the bottom panel is scored based on reciprocal attention with the red-circled token in the top panel. In both cases, the hand token in the opposite panel receives the strongest reciprocal attention, validating that RAVM successfully identifies semantically aligned token pairs for value mixing. This reinforces visual consistency without altering spatial composition.
    }
    \label{fig:mutual_attention}
\end{figure}

\section{Method}
\label{sec:method}

Our goal is to generate coherent storyboard panels from freeform text while preserving character identity across diverse compositions. An overview of our method is shown in Figure~\ref{fig:method_overview}. Given a natural language narrative, a large language model (LLM)--GPT-4o in our implementation--decomposes it into a shared reference panel prompt and $n$ scene-specific panel prompts, which are then jointly rendered using a pre-trained diffusion model. To ensure consistency without retraining, we introduce two complementary mechanisms, \textit{Latent Panel Anchoring (LPA)} and \textit{Reciprocal Attention Value Mixing (RAVM)}. LPA pairs each panel with a shared reference, thereby leveraging the model's self-attention mechanism to promote visual consistency between panels (\Cref{sec:latent_panel_anchoring}). While prompt-guided anchoring provides a useful bias, it alone is insufficient in scenes with complex layouts or ambiguous references. RAVM further enhances consistency by softly blending visual features between corresponding tokens across panels based on bidirectional attention cues (\Cref{sec:mutual-attention}). This blending preserves the expressive diversity of the model while reinforcing consistency with the reference panel. Together, LPA and RAVM enable coherent character synthesis and expressive scene composition without modifying the model or training procedure.

\subsection{Latent Panel Anchoring}
\label{sec:latent_panel_anchoring}

Our method generates a sequence of storyboard panels from a narrative text input, with consistent character identity and layout diversity across scenes. We begin by using an LLM~\cite{openai2024gpt4technicalreport} to decompose the input into a single reference prompt--which describes all recurring characters or objects--and a sequence of $n$ scene-level prompts, one per panel. These prompts are then paired as described below and fed into a pre-trained text-to-image diffusion model.

For each panel in the storyboard, we generate a composite prompt that combines a shared reference prompt with a scene-specific prompt. The reference prompt is designed to depict all recurring characters or objects in the story, while each scene prompt describes a distinct moment in the narrative. We structure each composite prompt as: “A storyboard of [reference prompt] (top) and [scene prompt] (bottom),” encouraging the model to depict a consistent character in varying contexts across stacked sub-panels.

Each composite prompt conditions the model to generate a two-part latent grid: the top half $R$ evolves into the reference sub-panel, and the bottom half $p_i$ into the target scene sub-panel, see \Cref{fig:method_overview}. We construct a batch of $n$ such latent grids, one for each scene, so that all target sub-panels are conditioned on the same reference prompt. To maintain consistency, we jointly denoise the batch and, after each transformer block, overwrite the (now different) top halves $R'_i$ of each latent, with that from the first batch item ($R'_1$). This ensures that every scene evolves relative to a shared, synchronized depiction of the main character(s). After generation, we discard the reference sub-panels and retain only the $n$ target sub-panels as the final storyboard.

As illustrated in Figure~\ref{fig:key-pca}, transformer-based diffusion models exhibit structured attention behavior: tokens corresponding to the same object--such as a character’s hair, clothing, or limbs--tend to form tight clusters in key-space. These internal “cliques” facilitate soft feature sharing between semantically aligned tokens, even when they are spatially distant in the image. This behavior was previously demonstrated in the previous UNet-based models~\cite{pnpDiffusion2022, geyer2023tokenflow, avrahami2024diffuhaul, tewel2024consistory}.

This consistency mechanism enables the model to propagate texture and style within and across panels. Latent Panel Anchoring leverages this emergent structure by placing a reference depiction of the character in every latent grid, allowing attention layers to align and blend visual features between the top and bottom sub-panels.

While prompt-guided anchoring and shared attention provide a strong inductive bias toward consistency, they are not always sufficient--particularly in scenes involving large pose variation, complex spatial layouts, or ambiguous references. To further reinforce token-level alignment across panels, we introduce a complementary mechanism in the next section.

\subsection{Reciprocal Attention Value Mixing (RAVM)}
\label{sec:mutual-attention}

While Latent Panel Anchoring encourages high-level visual consistency across storyboard panels, it may fail to preserve fine-grained identity--particularly when characters appear in different poses or spatial arrangements. \textit{Reciprocal Attention Value Mixing (RAVM)} addresses this by reinforcing cross-panel correspondences between semantically aligned tokens through soft feature blending.

In attention-based diffusion models, each token's representation is updated using three components: keys, queries, and values. Prior works \cite{tewel2023keylocked, kumarry2022customdiffusion} have established that keys and queries influence spatial layout and attention weighting, while values encode fine-grained visual detail such as texture, color, and appearance. RAVM operates solely on the value vectors--modifying a token’s appearance without affecting its layout--making it a natural mechanism for preserving identity while retaining scene diversity.

\begin{figure}[t!]
    \centering
    \includegraphics[width=\linewidth]{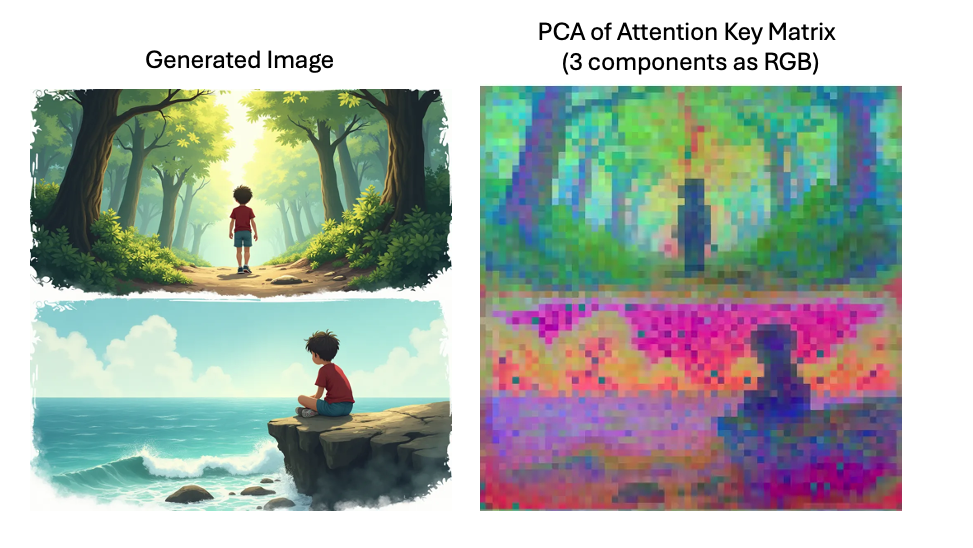}
    \caption{Semantic key clustering within a two-panel Flux-generated storyboard (left). We visualize the key vectors for each token at a mid-layer transformer block during diffusion step 12/28, by reducing them to three principal components, and displaying them as RGB values (right). Tokens corresponding to the character (e.g., hair, clothing) form tight clusters across panels, enabling consistent texture and style propagation via self-attention. In contrast, background tokens remain dispersed, reflecting limited cross-panel alignment.}
    \label{fig:key-pca}
\end{figure}

To decide \textit{which} value vectors to mix, we identify pairs of tokens that attend strongly to each other across stacked panels. These reciprocal relationships frequently emerge between semantically aligned regions--such as a character’s face or clothing--and are a key reason Latent Panel Anchoring works effectively (see Figure~\ref{fig:key-pca}). We make this structure explicit by interpreting the two-panel latent as a \textit{directed bipartite attention graph}: one set of nodes corresponds to tokens in the reference sub-panel, the other to tokens in the target sub-panel, and edge weights are given by attention values. We define a reciprocal attention score for each cross-panel token pair as the minimum of the attention in both directions, and selectively blend value vectors for those with the highest mutual connectivity.

This approach allows RAVM to softly propagate texture and style between corresponding regions, reinforcing visual consistency without overriding spatial variation or requiring explicit supervision.

Formally, let $x \in \mathbb{R}^{2P \times d}$ denote the concatenated tokens of the reference and target sub-panels. The model computes:

\begin{equation}
Q = x W_Q, \quad K = x W_K, \quad A = \text{softmax}\left(\frac{Q K^\top}{\sqrt{d_k}}\right) \in \mathbb{R}^{2P \times 2P},
\end{equation}

where $A[i,j]$ is the attention from token $i$ to token $j$. We extract the cross-panel blocks:
\begin{align}
A_{tb} &= A[1{:}P,\; P{:}2P], \\
A_{bt} &= A[P{:}2P,\; 1{:}P],
\end{align}
corresponding to top-to-bottom and bottom-to-top attention.

We define the \textit{reciprocal attention score} between a top token $u$ and bottom token $v$ as:
\begin{equation}
\operatorname{RA}(u, v) := \min(A_{tb}[u,v], A_{bt}[v,u]),
\end{equation}
yielding a symmetric matrix $M \in \mathbb{R}^{P \times P}$ of bidirectional scores, which during inference can be efficiently computed for all tokens by:
\begin{equation}
M := \min(A_{bt}, A_{tb}^T)\qquad\text{(elementwise)}
\end{equation}
We maintain an exponential moving average $\bar{M}$ of $M$ across transformer layers and diffusion steps.

To extract high-confidence correspondences, we analyze the weights on the edges crossing the bipartite graph cut. We apply Otsu’s thresholding method~\cite{otsu1975threshold} to the reciprocal attention matrix $\bar{M}$ and use morphological filtering to clean the resulting binary mask. For each selected bottom token $v$, we identify the top-panel token $u^*$ with the highest reciprocal score:
\begin{equation}
u^* = \arg\max_{u} \bar{M}[u, v],
\end{equation}
and apply a soft value update:
\begin{equation}
V'_v = \lambda V_v + (1 - \lambda) V_{u^*},
\end{equation}
where $V_i$ is the value vector of token $i$, and $\lambda$ is a mixing weight. Since keys and queries remain unchanged, the spatial layout and attention dynamics of the scene are preserved.

By reinforcing only the strongest reciprocal connections across the attention graph, RAVM enhances character consistency without suppressing scene diversity or altering the model’s generative flexibility.

\begin{figure}[t!]
  \centering
  \includegraphics[width=0.9\linewidth]{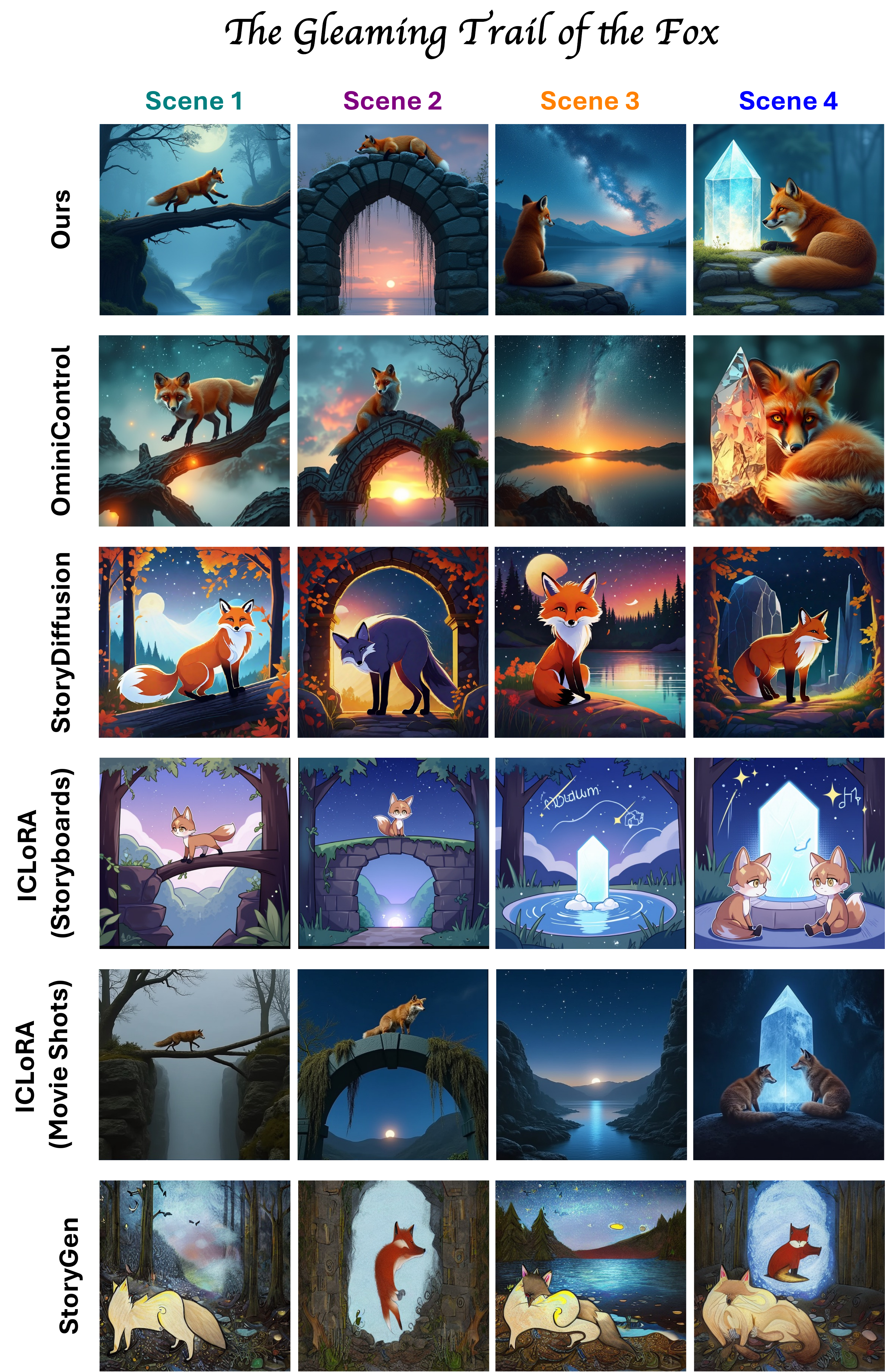}
  \caption{
  A four-panel storyboard featuring \textit{Blackpaw}, a shimmering fox of the ancient celestial forest. Each scene is grounded in a specific narrative beat from a longer story (full text in the supplementary). Our method preserves character consistency while supporting expressive spatial framing and richly atmospheric environments. Even as Blackpaw varies in pose, size, and placement across panels, the evolving backgrounds remain narratively grounded and visually coherent. For reference, the key visual moments are drawn from the following excerpts:
  \textcolor{teal}{“…With a flick of his glowing tail, he bounded across a fallen tree stretched precariously over a mist-shrouded ravine that gleamed faintly}
  \textcolor{violet}{…Perched atop a broken archway of ancient stone, vines and silver moss hanging around him, Blackpaw gazed out over the glowing forest as twilight deepened.}
  \textcolor{orange}{…From the edge of a luminous lake mirroring the heavens perfectly, he watched a meteor shower ignite the sky, each fiery streak mirrored twice over.}
  \textcolor{blue}{…Curling beside a pulsing crystal monolith, he dreamed…”}
  See the Appendix (Section~\ref{sec:appendix-full-story-texts}) for the full story text.
  }
\label{fig:storyboard-comparison}
\end{figure}

\section{Experiments}
\label{sec:experiments}

We evaluate our method both qualitatively and quantitatively, focusing on three core dimensions: prompt alignment, character consistency, and scene diversity. To support this, we introduce the \textbf{Rich Storyboard Benchmark}, designed to test narrative and compositional expressiveness beyond the scope of existing identity-focused datasets (see Appendix Section~\ref{sec:appendix-rich-story-benchmark}). We also evaluate on the DS-500 benchmark~\cite{he2024dreamstory} to demonstrate generalizability, (see Appendix Section~\ref{sec:appendix-ds-500} for results).

Section~\ref{sec:baselines} outlines the methods compared; Section~\ref{sec:benchmark} details our benchmark and metrics. Results are reported in Sections~\ref{sec:qualitative-eval} and~\ref{sec:quantitative-eval}, with human preference scores in Section~\ref{sec:user-study}.

\subsection{Baselines and Comparison Setup}
\label{sec:baselines}

\sloppy{We compare against story-centric models: \textbf{StoryDiffusion}~\cite{zhou2024StoryDiffusion}, which introduces Consistent Self-Attention; \textbf{IC-LoRA}~\cite{Huang2024InContextLF}, evaluated in storyboard and movie-shot finetuned variants; and \textbf{StoryGen}~\cite{liu2024storygen}, an autoregressive generator driven by scene prompts.
}
We also include \textbf{OminiControl}~\cite{tan2024ominicontrol}, which leverages a trained image encoder and a reference image to guide layout and style, outperforming prior encoder-based methods.

For ablations, we evaluate the \textbf{Flux base model} (no consistency mechanisms), and a version with \textbf{Latent Panel Anchoring (LPA)} only, isolating the contribution of Reciprocal Attention Value Mixing (RAVM). We also experiment with varying the value mixing coefficient~$\lambda$ to assess its effect on consistency and expressiveness.

\begin{figure*}[t]
    \centering
    \includegraphics[width=0.9\textwidth]{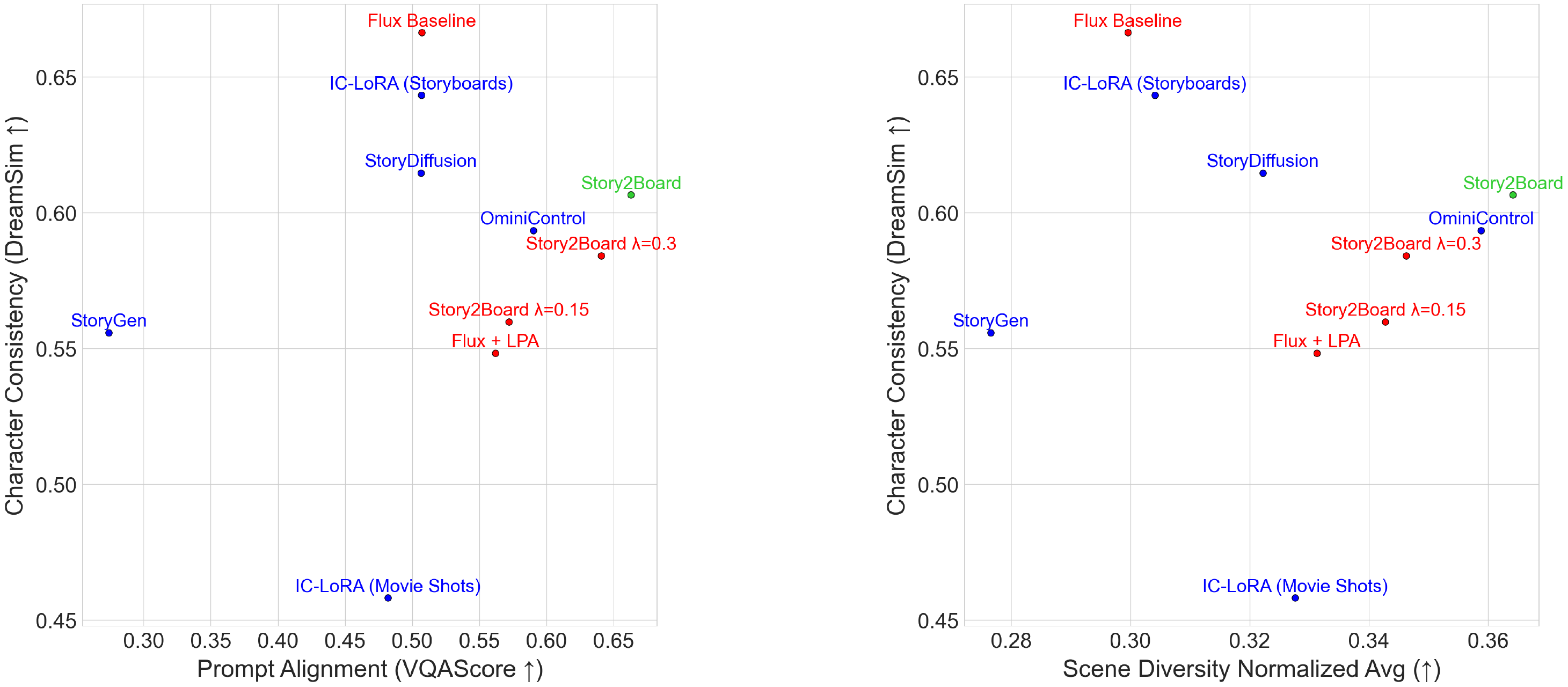}
    \caption{
    \textbf{Left: Character Consistency vs. Prompt Alignment.} \textit{Story2Board} achieves the best tradeoff, outperforming all baselines and ablations. Prompt alignment (x-axis) is measured via VQAScore and character consistency (y-axis) via DreamSim. The Flux baseline exhibits unusually high consistency due to its collapsed behavior--rendering similar characters across panels with minimal pose or appearance variation--yet struggles with prompt grounding.
    \textbf{Right: Scene Diversity vs. Character Consistency.} Our method maintains high identity fidelity while enabling significantly more layout variation than competing methods. Scene Diversity (x-axis) is our proposed metric (details in supplementary), while character consistency (y-axis) is again measured via DreamSim. Note that IC-LoRA baselines (Movie Shots and Storyboards) operate only on 4-panel sequences and are not applicable to longer formats.
    }
    \label{fig:metrics}
\end{figure*}

\subsection{Benchmark and Evaluation Metrics}
\label{sec:benchmark}

\paragraph{Rich Storyboard Benchmark.} 
Existing benchmarks focus primarily on identity preservation and do not reflect the compositional or cinematic demands of visual storytelling. To address this gap, we introduce the Rich Storyboard Benchmark, a set of 100 open-domain story prompts, each decomposed into seven richly detailed scene-level descriptions. The benchmark emphasizes dynamic layout, spatial diversity, and character-scene interaction--all critical for assessing visual narrative quality beyond identity fidelity.

\paragraph{Metrics.} 
We evaluate prompt alignment using VQAScore~\cite{Lin2024EvaluatingTG}, character consistency using DreamSim~\cite{Fu2023DreamSimLN}, and scene diversity using our novel metric, which quantifies variation in the character’s size, position, and pose across panels--capturing how flexibly the model composes the subject within the scene. Implementation details can be found in the Appendix (Section~\ref{sec:appendix-implementation}).

\subsection{Qualitative Evaluation}
\label{sec:qualitative-eval}

\paragraph{Full Storyboard Comparison.}

\begin{figure*}
  \centering
  \includegraphics[width=\linewidth]{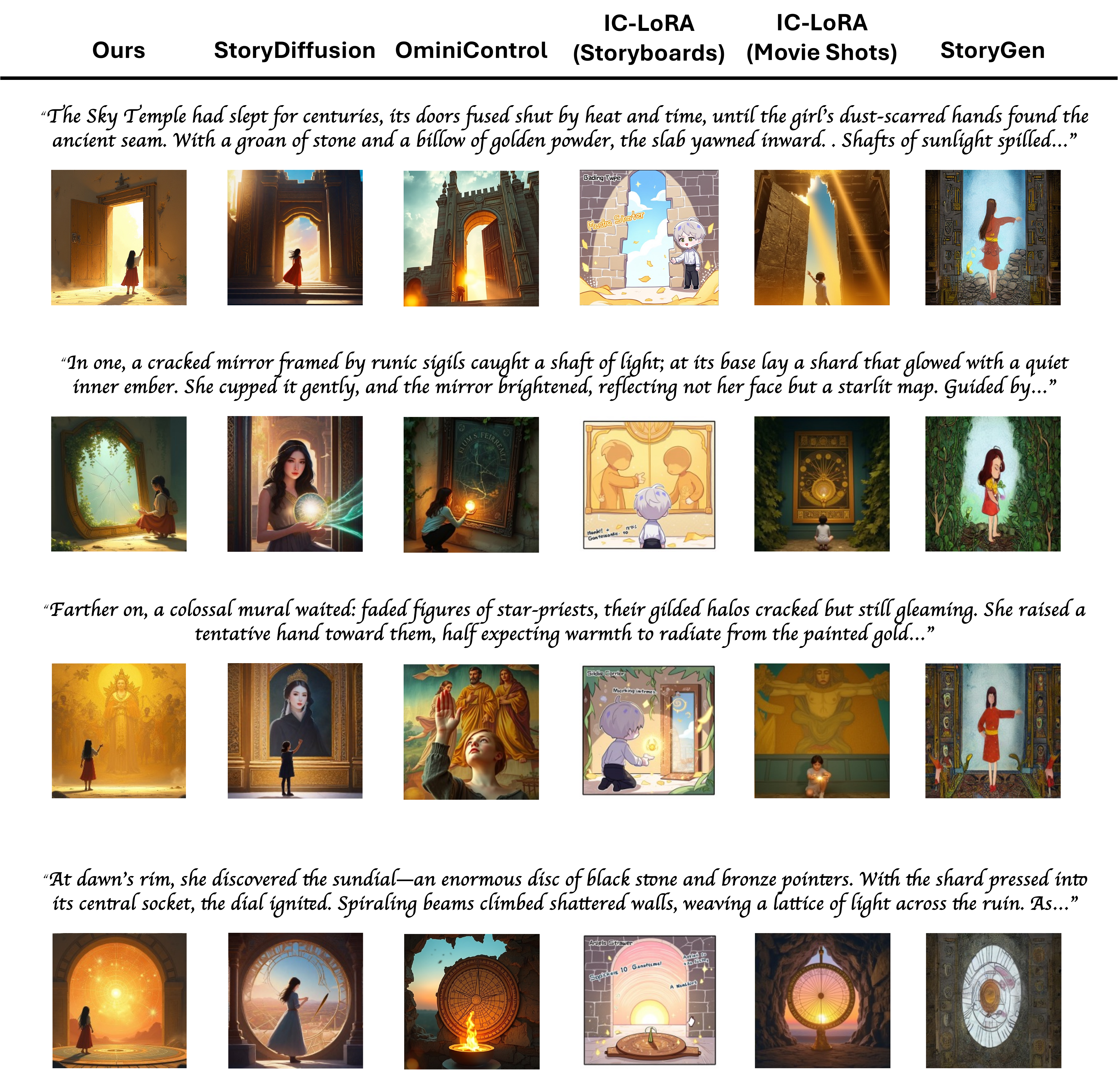}
  \caption{\textbf{Qualitative comparison of multi-panel storyboards.}
  Our method (\textsc{Story2Board}, left column) achieves a three-way balance:
  \emph{scene diversity} (varying viewpoints, scale, and richly grounded
  backgrounds), \emph{character consistency} (stable appearance and silhouette),
  and tight \emph{prompt alignment}.  Baseline systems each miss at least one of
  these axes: \textsc{StoryDiffusion} varies layouts but allows the heroine’s
  features to drift; \textsc{OminiControl} creates atmospheric backdrops yet
  occasionally omits the protagonist; \textsc{IC-LoRA Storyboards} fixes the
  camera and produces stylised cartoon frames, limiting narrative variety;
  \textsc{IC-LoRA Movie Shots} shows wider layouts but often mis-matches prompt
  details; \textsc{StoryGen} produces stylized frames, but struggles with narrative continuity and compositional coherence across panels.
  By maintaining identity while continuously re-contextualising the
  scene, \textsc{Story2Board} delivers the most faithful and visually engaging
  storyboard among current approaches.}
  \label{fig:full_story}
\end{figure*}

We present 4-panel storyboard sequences for two representative stories rendered by each method (Figures~\ref{fig:storyboard-comparison} and~\ref{fig:full_story}). Our method achieves a stronger balance across prompt alignment, character consistency, and scene diversity. It supports varied framing and character positioning while maintaining coherent, visually rich environments. Baselines tend to overfit one aspect: StoryDiffusion favors centered subjects; IC-LoRA repeats compositional templates; and OminiControl often omits off-center characters. Our method accommodates these challenges, yielding coherent, expressive storyboards.

\subsection{Quantitative Evaluation}
\label{sec:quantitative-eval}

We evaluate on both the Rich Storyboard Benchmark and DS-500~\cite{he2024dreamstory}. Metrics are computed per storyboard and averaged.

Figure~\ref{fig:metrics} shows that our method dominates the Pareto front in prompt alignment and character consistency. We visualize metrics pairwise to better expose tradeoffs: for instance, models with high character consistency often achieve it by sacrificing prompt alignment or layout flexibility. The Flux baseline exemplifies this pattern--it attains strong consistency scores by rendering nearly identical characters across panels, but lacks sensitivity to prompt-specific content, resulting in lower alignment and limited scene variation.

\begin{figure}
    \centering
    \includegraphics[width=0.9\linewidth]{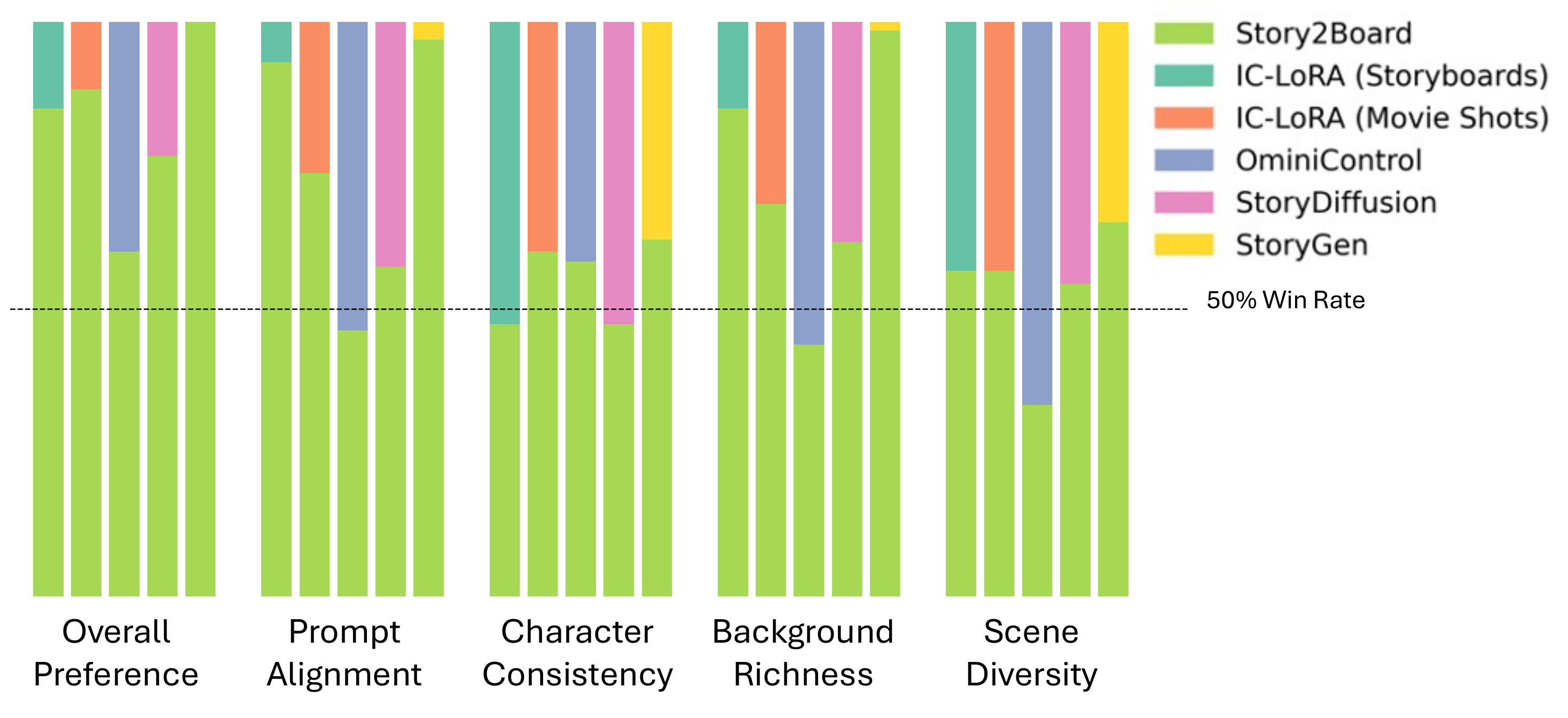}
    \caption{
    \textbf{User Study Results.} Participants compared Story2Board to competing systems across five evaluation dimensions. Our method is preferred overall, and achieves strong performance across all categories. While some baselines edge ahead in isolated metrics, Story2Board strikes the best balance between character consistency, visual richness, and narrative alignment--a key strength for storyboard generation.
    }
    \label{fig:user-study}
\end{figure}

\paragraph{DS-500 Evaluation.}
To assess generalization beyond our benchmark, we also evaluate on DS-500~\cite{he2025anystory}, a storyboard dataset with shorter prompts and minimal scene evolution. While not designed to test layout or narrative expressivity, DS-500 remains a useful baseline for identity coherence. Our method performs competitively on this benchmark, despite its focus on richer visual storytelling. Full results and comparisons are provided in the Appendix (Section~\ref{sec:appendix-ds-500}).

\paragraph{Ablation Study.}
All figures include two ablations of our method: a vanilla Flux baseline (no consistency mechanisms), and Flux with Latent Panel Anchoring (LPA) only. Our full method outperforms both, highlighting the complementary roles of LPA and Reciprocal Attention Value Mixing (RAVM). While LPA alone improves layout coherence, RAVM significantly boosts character consistency without harming layout diversity.

We further analyze the effect of the mixing parameter $\lambda$ in the RAVM update. Increasing $\lambda$ leads to a clear improvement in character consistency, as stronger blending amplifies the influence of semantically aligned reference tokens. Interestingly, we also observe gains in prompt alignment and scene diversity. While RAVM only modifies the value vectors--which capture texture and fine appearance details--we hypothesize that stabilizing the character's appearance helps the model more clearly separate and render background elements across panels, resulting in richer scenes and improved prompt fidelity.

\subsection{User Study}
\label{sec:user-study}

To supplement our quantitative evaluation, we conducted a large-scale user study via the Amazon Mechanical Turk (AMT) platform, using all 100 stories from our Rich Storyboard Benchmark. For each story, we generated 4-panel storyboards using our method and each of the competing baselines. Each worker task consisted of a pairwise comparison between two storyboards (one from our method and one from a baseline), with each comparison focused on one of five criteria: overall preference, prompt alignment, character consistency, background richness, and scene diversity. In total, 500 such tasks were created, and each was completed by three independent workers. \Cref{fig:user-study} summarizes the results, showing in green the preference rate of our method over each baseline for each of the five criteria.

Our method was the most preferred overall, winning the majority of pairwise comparisons in the “Overall Preference” category. This suggests that when users evaluated storyboards holistically, they consistently favored our approach over all baselines.

However, the results reveal more nuanced trade-offs in some of the other dimensions. OminiControl outperformed our method in prompt alignment, background richness, and scene diversity--likely due to its strong compositional control via encoder-based conditioning. IC-LoRA (Storyboards) and StoryDiffusion were preferred for character consistency, reflecting their explicit focus on preserving visual identity across frames. In contrast, our method favors soft guidance mechanisms that support dynamic layouts and flexible character framing, which may explain its advantage in overall appeal despite lower scores in some of the narrower aspects.

\subsection{Limitations}
\label{sec:limitations}

\begin{figure}
    \centering
    \includegraphics[width=\linewidth]{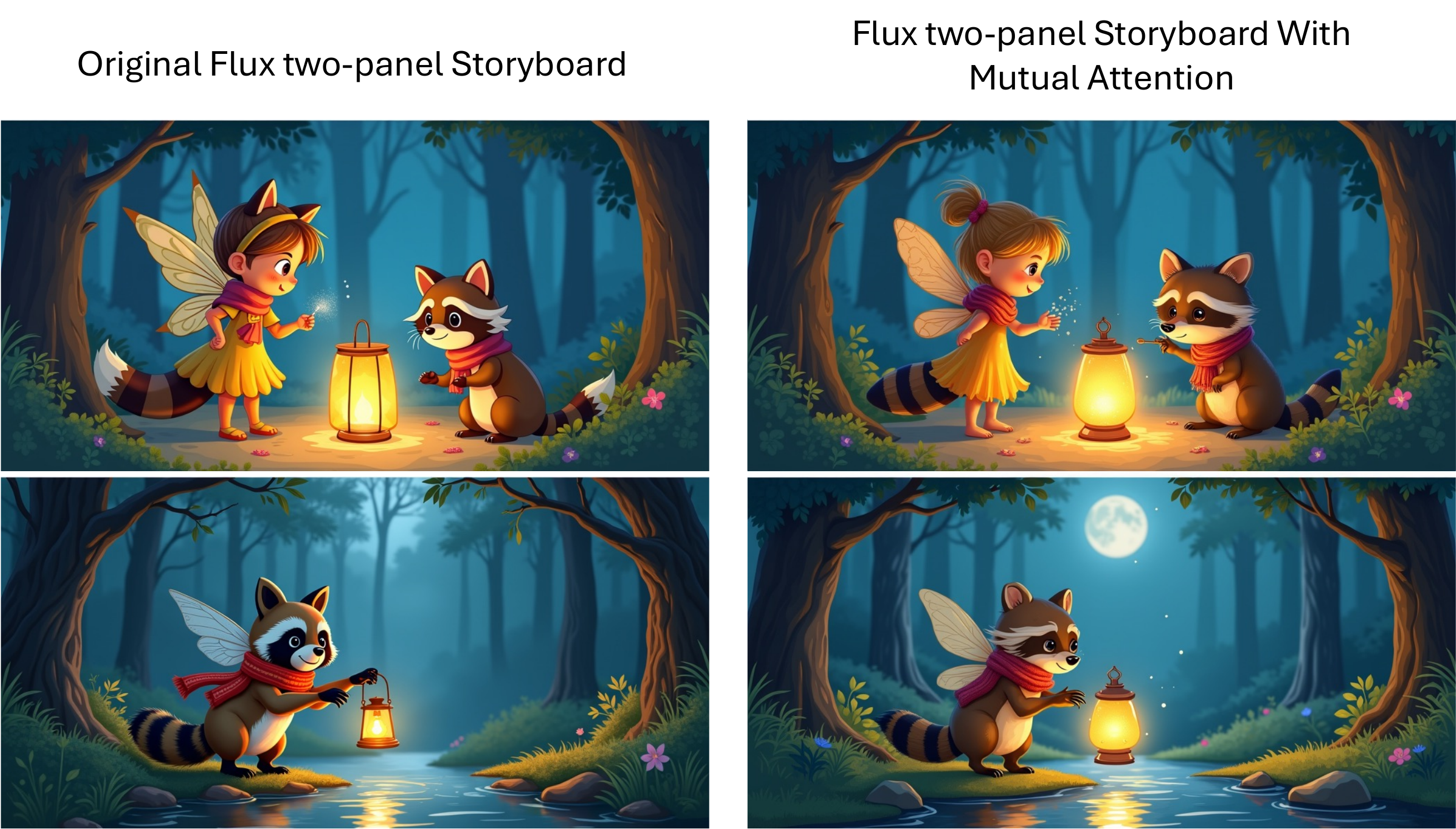}
    \caption{
    \textbf{Attention Entanglement in Flux.} Left: A two-panel storyboard generated by Flux without our method. Attention entanglement causes the fairy to erroneously inherit the raccoon's tail in the top panel, while in the bottom panel the raccoon adopts the fairy's wings. Right: With our Mutual Attention (MA) mechanism, these misattributions persist, but their visual appearance becomes consistent across panels. MA also improves the consistency of other visual elements--such as the raccoon’s tail and the lantern--demonstrating the broader stabilizing influence of token-level value mixing. When entangled representations are already present in the base model, our method propagates rather than corrects them.
    }
    \label{fig:limitations}
\end{figure}

Our method builds directly on the internal attention dynamics of modern text-to-image diffusion models such as Flux and Stable Diffusion. While this enables us to enhance cross-panel consistency without additional training or architectural changes, it also means we inherit certain limitations of the underlying models. One well-documented issue is \textit{attention entanglement}--including phenomena such as incorrect attribute binding, object fusion, and semantic misassignment--where separate concepts or entities interfere with one another during generation~\cite{dahary2024boundedattention}. Since our approach reinforces token-level correspondences based on mutual attention, it may inadvertently propagate these entanglements if they persist across the denoising trajectory. Importantly, our method does not exacerbate these effects; it simply cannot prevent them when they are already embedded in the model’s native attention structure. An example of such entanglement is shown in Figure~\ref{fig:limitations}.

\section{Summary and Discussion}
\label{sec:discussion}

We introduced Story2Board, a training-free framework for generating visually consistent and compositionally rich storyboards from text. By leveraging reciprocal attention patterns between tokens across panels, our method reinforces character identity while preserving layout diversity. Through extensive experiments on our proposed benchmark and a standard existing dataset, we demonstrate that Story2Board enables more dynamic, expressive visual storytelling than prior approaches.

\begin{figure*}
  \centering
  \includegraphics[width=\textwidth]{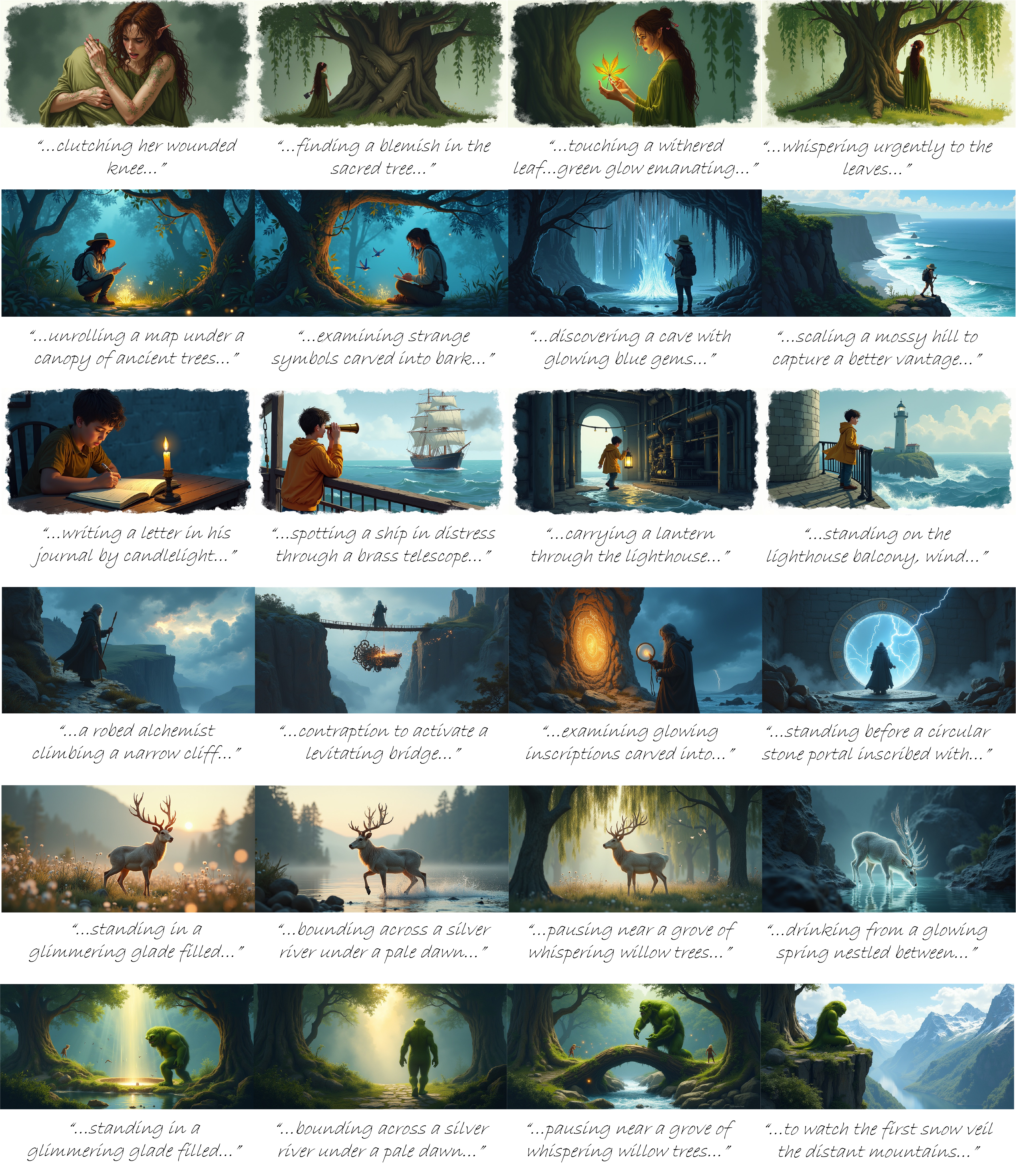}
  \caption{Additional storyboards generated by our method.}
  \label{fig:supp_storyboards_A}
\end{figure*}

\begin{figure*}[t!]
    \centering
    \includegraphics[width=0.89\linewidth]{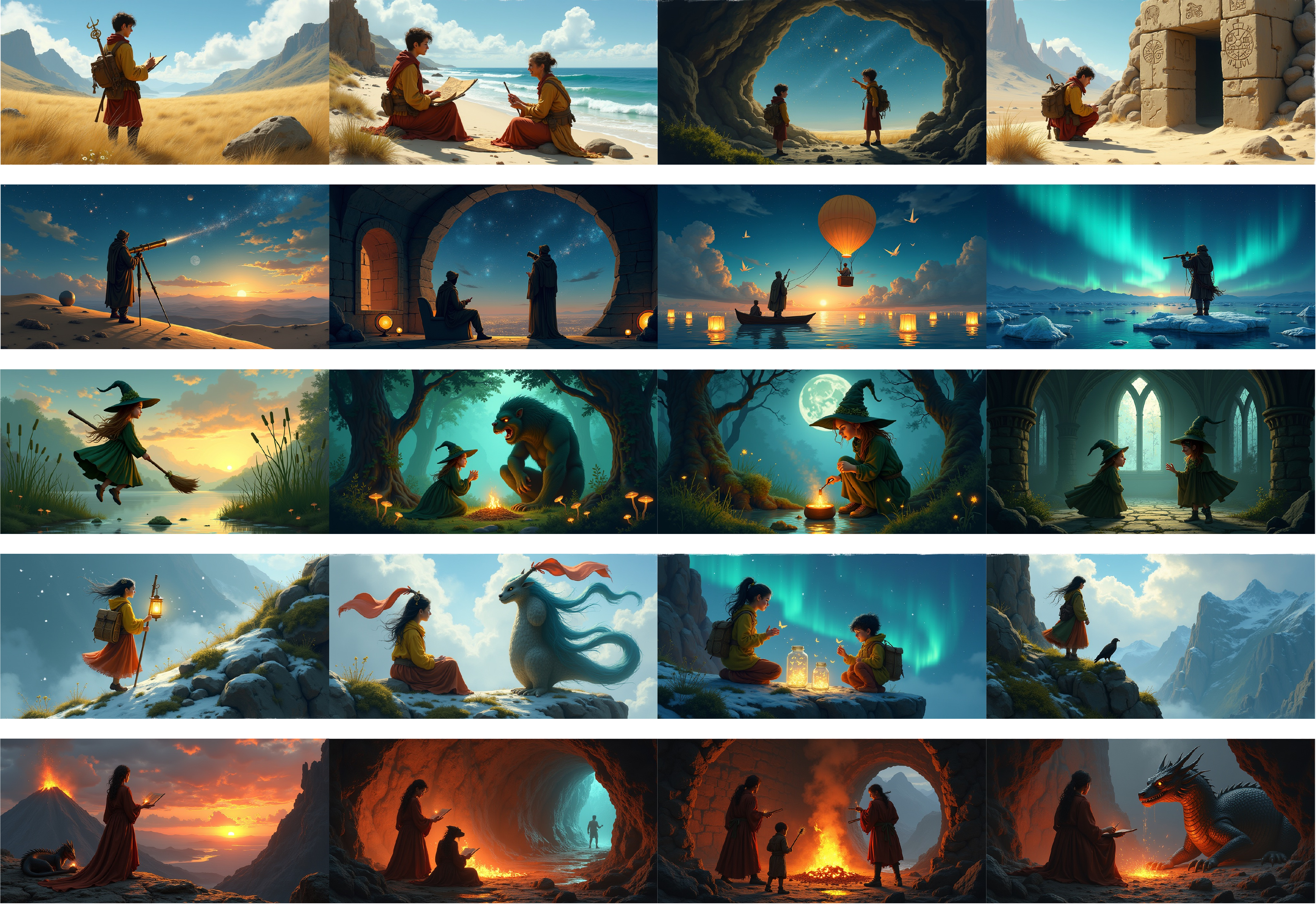}
    \caption{
    \textbf{Characters and Strangers.} Each row shows a storyboard from the Rich Storyboard Benchmark where the main character encounters unfamiliar figures, testing the model's ability to maintain character identity while integrating diverse background elements. Additional storyboards and prompts for each panel are provided in the supplementary material.}
    \label{fig:charcters_and_strangers}
\end{figure*}

\begin{figure*}
  \centering
  \includegraphics[width=\textwidth]{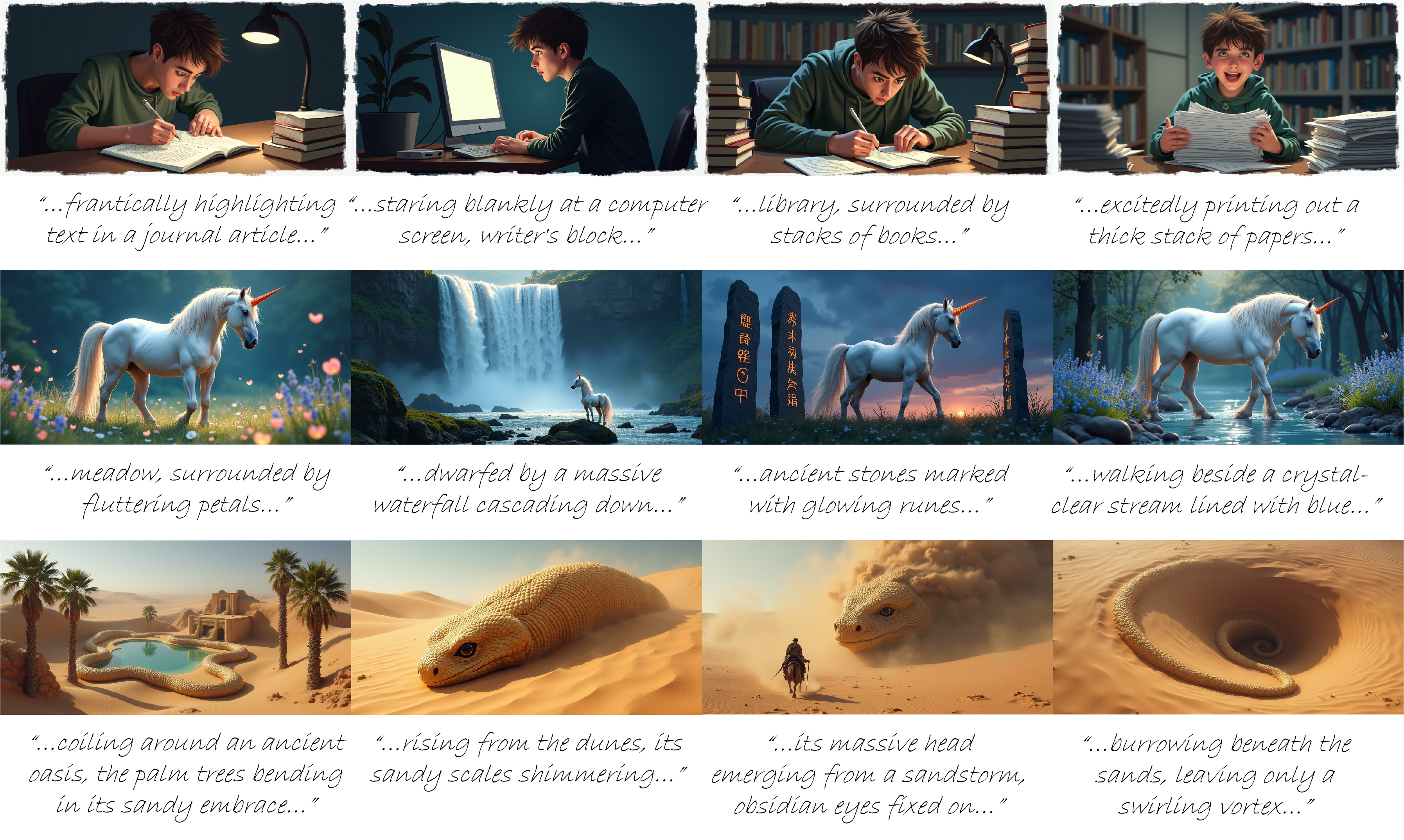}
  \caption{Additional storyboards generated by our method.}
  \label{fig:supp_storyboards_B}
\end{figure*}

\clearpage
\flushend
\clearpage
\flushend
\bibliographystyle{plain}
\bibliography{paletteDiff}

\begin{thebibliography}{10}

\bibitem{amazonmturk}
{Amazon Mechanical Turk}.
\newblock Amazon mechanical turk.
\newblock \url{https://www.mturk.com/}, 2025.
\newblock Accessed: 2025-05-20.

\bibitem{animatorisland_breathingroom}
{Animator Island}.
\newblock Composition: What is breathing room?
\newblock \url{https://www.animatorisland.com/composition-what-is-breathing-room/}, 2014.
\newblock Accessed: 2025-05-12.

\bibitem{avrahami2024diffuhaul}
Omri Avrahami, Rinon Gal, Gal Chechik, Ohad Fried, Dani Lischinski, Arash Vahdat, and Weili Nie.
\newblock Diffuhaul: A training-free method for object dragging in images.
\newblock In {\em SIGGRAPH Asia 2024 Conference Papers}, SA '24, New York, NY, USA, 2024. Association for Computing Machinery.

\bibitem{avrahami2024chosen}
Omri Avrahami, Amir Hertz, Yael Vinker, Moab Arar, Shlomi Fruchter, Ohad Fried, Daniel Cohen-Or, and Dani Lischinski.
\newblock The chosen one: Consistent characters in text-to-image diffusion models.
\newblock In {\em ACM SIGGRAPH 2024 Conference Papers}, SIGGRAPH '24, New York, NY, USA, 2024. Association for Computing Machinery.

\bibitem{flux}
{Black Forest Labs}.
\newblock Flux.
\newblock \url{https://github.com/black-forest-labs/flux}, 2024.

\bibitem{block_visualstory}
Bruce Block.
\newblock {\em The Visual Story: Creating the Visual Structure of Film, TV, and Digital Media}.
\newblock Focal Press, 3rd edition, 2020.

\bibitem{dahary2024boundedattention}
Omer Dahary, Or~Patashnik, Kfir Aberman, and Daniel Cohen-Or.
\newblock Be yourself: Bounded attention for multi-subject text-to-image generation.
\newblock In {\em European Conference on Computer Vision}, pages 432--448. Springer, 2024.

\bibitem{Esser2024ScalingRF}
Patrick Esser, Sumith Kulal, A.~Blattmann, Rahim Entezari, Jonas Muller, Harry Saini, Yam Levi, Dominik Lorenz, Axel Sauer, Frederic Boesel, Dustin Podell, Tim Dockhorn, Zion English, Kyle Lacey, Alex Goodwin, Yannik Marek, and Robin Rombach.
\newblock Scaling rectified flow transformers for high-resolution image synthesis.
\newblock {\em ArXiv}, abs/2403.03206, 2024.

\bibitem{filmmakersacademy_negspace}
{Filmmakers Academy}.
\newblock Negative space: Film composition guide.
\newblock \url{https://www.filmmakersacademy.com/blog-negative-space-film/}, 2025.
\newblock Accessed: 2025-05-12.

\bibitem{Fu2023DreamSimLN}
Stephanie Fu, Netanel~Y. Tamir, Shobhita Sundaram, Lucy Chai, Richard Zhang, Tali Dekel, and Phillip Isola.
\newblock Dreamsim: Learning new dimensions of human visual similarity using synthetic data.
\newblock {\em ArXiv}, abs/2306.09344, 2023.

\bibitem{geyer2023tokenflow}
Michal Geyer, Omer Bar-Tal, Shai Bagon, and Tali Dekel.
\newblock Tokenflow: Consistent diffusion features for consistent video editing.
\newblock {\em arXiv preprint arXiv:2307.10373}, 2023.

\bibitem{he2024dreamstory}
Huiguo He, Huan Yang, Zixi Tuo, Yuan Zhou, Qiuyue Wang, Yuhang Zhang, Zeyu Liu, Wenhao Huang, Hongyang Chao, and Jian Yin.
\newblock Dreamstory: Open-domain story visualization by llm-guided multi-subject consistent diffusion.
\newblock {\em arXiv preprint arXiv:2407.12899}, 2024.

\bibitem{he2025anystory}
Junjie He, Yuxiang Tuo, Binghui Chen, Chongyang Zhong, Yifeng Geng, and Liefeng Bo.
\newblock Anystory: Towards unified single and multiple subject personalization in text-to-image generation.
\newblock {\em arXiv preprint arXiv:2501.09503}, 2025.

\bibitem{ho2020denoising}
Jonathan Ho, Ajay Jain, and Pieter Abbeel.
\newblock Denoising diffusion probabilistic models.
\newblock In {\em Proc.~NeurIPS}, 2020.

\bibitem{Huang2024InContextLF}
Lianghua Huang, Wei Wang, Zhigang Wu, Yupeng Shi, Huanzhang Dou, Chen Liang, Yutong Feng, Yu~Liu, and Jingren Zhou.
\newblock In-context lora for diffusion transformers.
\newblock {\em ArXiv}, abs/2410.23775, 2024.

\bibitem{OpenCLIP}
Gabriel Ilharco, Mitchell Wortsman, Ross Wightman, Cade Gordon, Nicholas Carlini, Rohan Taori, Achal Dave, Vaishaal Shankar, Hongseok Namkoong, John Miller, Hannaneh Hajishirzi, Ali Farhadi, and Ludwig Schmidt.
\newblock {OpenCLIP}, July 2021.

\bibitem{kirillov2023segment}
Alexander Kirillov, Eric Mintun, Nikhila Ravi, Hanzi Mao, Chloe Rolland, Laura Gustafson, Tete Xiao, Spencer Whitehead, Alexander~C. Berg, Wan-Yen Lo, Piotr Doll\'{a}r, and Ross Girshick.
\newblock Segment anything, 2023.

\bibitem{kumarry2022customdiffusion}
Nupur Kumari, Bingliang Zhang, Richard Zhang, Eli Shechtman, and Jun-Yan Zhu.
\newblock Multi-concept customization of text-to-image diffusion.
\newblock In {\em Proceedings of the IEEE/CVF Conference on Computer Vision and Pattern Recognition}, pages 1931--1941, 2023.

\bibitem{Lin2024EvaluatingTG}
Zhiqiu Lin, Deepak Pathak, Baiqi Li, Jiayao Li, Xide Xia, Graham Neubig, Pengchuan Zhang, and Deva Ramanan.
\newblock Evaluating text-to-visual generation with image-to-text generation.
\newblock In {\em European Conference on Computer Vision}, 2024.

\bibitem{liu2024storygen}
Chang Liu, Haoning Wu, Yujie Zhong, Xiaoyun Zhang, Yanfeng Wang, and Weidi Xie.
\newblock Intelligent grimm -- open-ended visual storytelling via latent diffusion models, 2024.

\bibitem{liu2023video}
Shaoteng Liu, Yuechen Zhang, Wenbo Li, Zhe Lin, and Jiaya Jia.
\newblock Video-p2p: Video editing with cross-attention control.
\newblock {\em arXiv preprint arXiv:2303.04761}, 2023.

\bibitem{openai2024gpt4technicalreport}
OpenAI, Josh Achiam, Steven Adler, Sandhini Agarwal, Lama Ahmad, Ilge Akkaya, Florencia~Leoni Aleman, Diogo Almeida, Janko Altenschmidt, Sam Altman, Shyamal Anadkat, Red Avila, Igor Babuschkin, Suchir Balaji, Valerie Balcom, Paul Baltescu, Haiming Bao, Mohammad Bavarian, Jeff Belgum, Irwan Bello, Jake Berdine, Gabriel Bernadett-Shapiro, Christopher Berner, Lenny Bogdonoff, Oleg Boiko, Madelaine Boyd, Anna-Luisa Brakman, Greg Brockman, Tim Brooks, Miles Brundage, Kevin Button, Trevor Cai, Rosie Campbell, Andrew Cann, Brittany Carey, Chelsea Carlson, Rory Carmichael, Brooke Chan, Che Chang, Fotis Chantzis, Derek Chen, Sully Chen, Ruby Chen, Jason Chen, Mark Chen, Ben Chess, Chester Cho, Casey Chu, Hyung~Won Chung, Dave Cummings, Jeremiah Currier, Yunxing Dai, Cory Decareaux, Thomas Degry, Noah Deutsch, Damien Deville, Arka Dhar, David Dohan, Steve Dowling, Sheila Dunning, Adrien Ecoffet, Atty Eleti, Tyna Eloundou, David Farhi, Liam Fedus, Niko Felix, Simón~Posada Fishman, Juston Forte, Isabella Fulford, Leo
  Gao, Elie Georges, Christian Gibson, Vik Goel, Tarun Gogineni, Gabriel Goh, Rapha Gontijo-Lopes, Jonathan Gordon, Morgan Grafstein, Scott Gray, Ryan Greene, Joshua Gross, Shixiang~Shane Gu, Yufei Guo, Chris Hallacy, Jesse Han, Jeff Harris, Yuchen He, Mike Heaton, Johannes Heidecke, Chris Hesse, Alan Hickey, Wade Hickey, Peter Hoeschele, Brandon Houghton, Kenny Hsu, Shengli Hu, Xin Hu, Joost Huizinga, Shantanu Jain, Shawn Jain, Joanne Jang, Angela Jiang, Roger Jiang, Haozhun Jin, Denny Jin, Shino Jomoto, Billie Jonn, Heewoo Jun, Tomer Kaftan, Łukasz Kaiser, Ali Kamali, Ingmar Kanitscheider, Nitish~Shirish Keskar, Tabarak Khan, Logan Kilpatrick, Jong~Wook Kim, Christina Kim, Yongjik Kim, Jan~Hendrik Kirchner, Jamie Kiros, Matt Knight, Daniel Kokotajlo, Łukasz Kondraciuk, Andrew Kondrich, Aris Konstantinidis, Kyle Kosic, Gretchen Krueger, Vishal Kuo, Michael Lampe, Ikai Lan, Teddy Lee, Jan Leike, Jade Leung, Daniel Levy, Chak~Ming Li, Rachel Lim, Molly Lin, Stephanie Lin, Mateusz Litwin, Theresa Lopez, Ryan
  Lowe, Patricia Lue, Anna Makanju, Kim Malfacini, Sam Manning, Todor Markov, Yaniv Markovski, Bianca Martin, Katie Mayer, Andrew Mayne, Bob McGrew, Scott~Mayer McKinney, Christine McLeavey, Paul McMillan, Jake McNeil, David Medina, Aalok Mehta, Jacob Menick, Luke Metz, Andrey Mishchenko, Pamela Mishkin, Vinnie Monaco, Evan Morikawa, Daniel Mossing, Tong Mu, Mira Murati, Oleg Murk, David Mély, Ashvin Nair, Reiichiro Nakano, Rajeev Nayak, Arvind Neelakantan, Richard Ngo, Hyeonwoo Noh, Long Ouyang, Cullen O'Keefe, Jakub Pachocki, Alex Paino, Joe Palermo, Ashley Pantuliano, Giambattista Parascandolo, Joel Parish, Emy Parparita, Alex Passos, Mikhail Pavlov, Andrew Peng, Adam Perelman, Filipe de~Avila Belbute~Peres, Michael Petrov, Henrique~Ponde de~Oliveira~Pinto, Michael, Pokorny, Michelle Pokrass, Vitchyr~H. Pong, Tolly Powell, Alethea Power, Boris Power, Elizabeth Proehl, Raul Puri, Alec Radford, Jack Rae, Aditya Ramesh, Cameron Raymond, Francis Real, Kendra Rimbach, Carl Ross, Bob Rotsted, Henri Roussez,
  Nick Ryder, Mario Saltarelli, Ted Sanders, Shibani Santurkar, Girish Sastry, Heather Schmidt, David Schnurr, John Schulman, Daniel Selsam, Kyla Sheppard, Toki Sherbakov, Jessica Shieh, Sarah Shoker, Pranav Shyam, Szymon Sidor, Eric Sigler, Maddie Simens, Jordan Sitkin, Katarina Slama, Ian Sohl, Benjamin Sokolowsky, Yang Song, Natalie Staudacher, Felipe~Petroski Such, Natalie Summers, Ilya Sutskever, Jie Tang, Nikolas Tezak, Madeleine~B. Thompson, Phil Tillet, Amin Tootoonchian, Elizabeth Tseng, Preston Tuggle, Nick Turley, Jerry Tworek, Juan Felipe~Cerón Uribe, Andrea Vallone, Arun Vijayvergiya, Chelsea Voss, Carroll Wainwright, Justin~Jay Wang, Alvin Wang, Ben Wang, Jonathan Ward, Jason Wei, CJ~Weinmann, Akila Welihinda, Peter Welinder, Jiayi Weng, Lilian Weng, Matt Wiethoff, Dave Willner, Clemens Winter, Samuel Wolrich, Hannah Wong, Lauren Workman, Sherwin Wu, Jeff Wu, Michael Wu, Kai Xiao, Tao Xu, Sarah Yoo, Kevin Yu, Qiming Yuan, Wojciech Zaremba, Rowan Zellers, Chong Zhang, Marvin Zhang, Shengjia
  Zhao, Tianhao Zheng, Juntang Zhuang, William Zhuk, and Barret Zoph.
\newblock Gpt-4 technical report, 2024.

\bibitem{Oquab2023DINOv2LR}
Maxime Oquab, Timoth{\'e}e Darcet, Th{\'e}o Moutakanni, Huy~Q. Vo, Marc Szafraniec, Vasil Khalidov, Pierre Fernandez, Daniel Haziza, Francisco Massa, Alaaeldin El-Nouby, Mahmoud Assran, Nicolas Ballas, Wojciech Galuba, Russ Howes, Po-Yao~(Bernie) Huang, Shang-Wen Li, Ishan Misra, Michael~G. Rabbat, Vasu Sharma, Gabriel Synnaeve, Huijiao Xu, Herv{\'e} J{\'e}gou, Julien Mairal, Patrick Labatut, Armand Joulin, and Piotr Bojanowski.
\newblock {DINOv2}: Learning robust visual features without supervision.
\newblock {\em ArXiv}, abs/2304.07193, 2023.

\bibitem{otsu1975threshold}
Nobuyuki Otsu et~al.
\newblock A threshold selection method from gray-level histograms.
\newblock {\em Automatica}, 11(285-296):23--27, 1975.

\bibitem{Podell2023SDXLIL}
Dustin Podell, Zion English, Kyle Lacey, A.~Blattmann, Tim Dockhorn, Jonas Muller, Joe Penna, and Robin Rombach.
\newblock {SDXL}: Improving latent diffusion models for high-resolution image synthesis.
\newblock {\em ArXiv}, abs/2307.01952, 2023.

\bibitem{ramesh2022hierarchical}
Aditya Ramesh, Prafulla Dhariwal, Alex Nichol, Casey Chu, and Mark Chen.
\newblock Hierarchical text-conditional image generation with {CLIP} latents.
\newblock {\em arXiv preprint arXiv:2204.06125}, 2022.

\bibitem{Rombach2021HighResolutionIS}
Robin Rombach, A.~Blattmann, Dominik Lorenz, Patrick Esser, and Bj{\"o}rn Ommer.
\newblock High-resolution image synthesis with latent diffusion models.
\newblock {\em 2022 IEEE/CVF Conference on Computer Vision and Pattern Recognition (CVPR)}, pages 10674--10685, 2021.

\bibitem{Saharia2022PhotorealisticTD}
Chitwan Saharia, William Chan, Saurabh Saxena, Lala Li, Jay Whang, Emily~L Denton, Kamyar Ghasemipour, Raphael Gontijo~Lopes, Burcu Karagol~Ayan, Tim Salimans, et~al.
\newblock Photorealistic text-to-image diffusion models with deep language understanding.
\newblock {\em Advances in Neural Information Processing Systems}, 35:36479--36494, 2022.

\bibitem{stablediffusion3}
{Stability AI}.
\newblock Stable diffusion 3: Next-generation text-to-image generation.
\newblock \\url{https://stability.ai/news/stable-diffusion-3}, 2024.

\bibitem{tan2024ominicontrol}
Zhenxiong Tan, Songhua Liu, Xingyi Yang, Qiaochu Xue, and Xinchao Wang.
\newblock Ominicontrol: Minimal and universal control for diffusion transformer.
\newblock {\em arXiv preprint arXiv:2411.15098}, 2024.

\bibitem{tewel2023keylocked}
Yoad Tewel, Rinon Gal, Gal Chechik, and Yuval Atzmon.
\newblock Key-locked rank one editing for text-to-image personalization.
\newblock In {\em ACM SIGGRAPH 2023 Conference Proceedings}, SIGGRAPH '23, 2023.

\bibitem{tewel2024consistory}
Yoad Tewel, Omri Kaduri, Rinon Gal, Yoni Kasten, Lior Wolf, Gal Chechik, and Yuval Atzmon.
\newblock Training-free consistent text-to-image generation.
\newblock {\em ACM Transactions on Graphics (TOG)}, 43(4):1--18, 2024.

\bibitem{Tewel2024TrainingFreeCT}
Yoad Tewel, Omri Kaduri, Rinon Gal, Yoni Kasten, Lior Wolf, Gal Chechik, and Yuval Atzmon.
\newblock Training-free consistent text-to-image generation.
\newblock {\em ArXiv}, abs/2402.03286, 2024.

\bibitem{pnpDiffusion2022}
Narek Tumanyan, Michal Geyer, Shai Bagon, and Tali Dekel.
\newblock Plug-and-play diffusion features for text-driven image-to-image translation.
\newblock In {\em Proceedings of the IEEE/CVF Conference on Computer Vision and Pattern Recognition}, pages 1921--1930, 2023.

\bibitem{vaswani2017attention}
Ashish Vaswani, Noam Shazeer, Niki Parmar, Jakob Uszkoreit, Llion Jones, Aidan~N Gomez, {\L}ukasz Kaiser, and Illia Polosukhin.
\newblock Attention is all you need.
\newblock {\em Advances in neural information processing systems}, 30, 2017.

\bibitem{Wei2023ELITEEV}
Yuxiang Wei, Yabo Zhang, Zhilong Ji, Jinfeng Bai, Lei Zhang, and Wangmeng Zuo.
\newblock {ELITE}: Encoding visual concepts into textual embeddings for customized text-to-image generation.
\newblock {\em ArXiv}, abs/2302.13848, 2023.

\bibitem{yang2024seed}
Shuai Yang, Yuying Ge, Yang Li, Yukang Chen, Yixiao Ge, Ying Shan, and Yingcong Chen.
\newblock Seed-story: Multimodal long story generation with large language model.
\newblock {\em arXiv preprint arXiv:2407.08683}, 2024.

\bibitem{ye2023ip-adapter}
Hu~Ye, Jun Zhang, Sibo Liu, Xiao Han, and Wei Yang.
\newblock {IP-Adapter}: Text compatible image prompt adapter for text-to-image diffusion models.
\newblock {\em arXiv}, abs/2308.06721, 2023.

\bibitem{Zhang2023MagicBrush}
Kai Zhang, Lingbo Mo, Wenhu Chen, Huan Sun, and Yu~Su.
\newblock Magicbrush: A manually annotated dataset for instruction-guided image editing.
\newblock In {\em Advances in Neural Information Processing Systems}, 2023.

\bibitem{zhou2024StoryDiffusion}
Yupeng Zhou, Daquan Zhou, Ming-Ming Cheng, Jiashi Feng, and Qibin Hou.
\newblock Storydiffusion: Consistent self-attention for long-range image and video generation.
\newblock {\em ArXiv}, abs/2405.01434, 2024.

\end{thebibliography}

\clearpage

\appendix

\renewcommand{\thetable}{\Alph{section}.\arabic{table}}
\renewcommand{\thefigure}{\Alph{section}.\arabic{figure}}
\setcounter{table}{0}
\setcounter{figure}{0}

\section{Appendix}

\subsection{Rich Storyboard Benchmark}
\label{sec:appendix-rich-story-benchmark}

To evaluate expressive visual storytelling, we introduce the \textbf{Rich Storyboard Benchmark}, a collection of 100 richly detailed and narratively structured story prompts. Each entry consists of a short, original story accompanied by a 7-panel storyboard breakdown. Prompts were generated using GPT-4o~\cite{openai2024gpt4technicalreport}, guided by instructions emphasizing narrative progression, visual diversity, and character continuity.

Unlike prior benchmarks that rely on templated prompts or static characters, our benchmark features dynamic compositions: characters move through environments that change over time, with evolving lighting, weather, architecture, and emotional tone.

\paragraph{GPT-4o Prompt (used to generate benchmark entries)}
\begin{quote}
You are a storyboard designer helping to create cinematic, visually expressive storyboards. Your task is to:

(1) Invent a short fantasy or human-centric story with a clear and exciting story arc.

(2) Make sure the story contains not only a memorable character, but also richly described background elements that evolve across scenes. These might include weather, lighting, ruins, architecture, mountains, oceans, caves, temples, streets, or crowds. Do not focus only on the character.

(3) Break down the story into seven distinct storyboard scene descriptions. Each should describe one key visual moment--something worth illustrating in a single panel. Replace names with a description of the character (e.g. instead of "Tom went to the store", say "A man went to the store").

Return:

- The full story

- A list of 7 storyboard panel prompts that visually represent the story
\end{quote}

\paragraph{Example Output (from GPT-4o)}
\begin{quote}
\item{
\textbf{Story:}
    Elias, the lone sailor in a battered wool cap and dark coat, stood calmly
    at the bow of his wooden ship, its prow cutting through the golden morning fog.
    The endless sea stretched around him, a liquid desert beneath a flickering sky.
    Later, Elias sat on a weathered crate beneath heavy coils of rope, gulls circling
    overhead, the ship creaking with age. He gazed outward, where the vastness made
    even the towering masts seem like toys adrift in a dream.
    At dusk, he leaned against the rail, lanterns swinging from the rigging above
    him as he stared at the glowing horizon, where the sky burned into the water.
    A massive arch of stone rose from the ocean ahead, so gigantic that Elias felt
    as small as a pebble beside it, looking up in awe.
    When night fell, he rested beside the ship's wheel, moonlight painting silver
    paths across the undulating waves. Eventually, only a quiet harbor awaited him,
    ropes swaying in the wind, the ship itself nowhere to be seen--as if it had drifted
    into legend.
}
\end{quote}

\textbf{Storyboard panel prompts:}

\begin{enumerate}
  \item A lone sailor in a wool cap and dark coat standing calmly at the bow of a wooden ship in golden morning fog
  \item A lone sailor in a wool cap and dark coat sitting on a crate beneath coiled ropes as gulls circle overhead
  \item A lone sailor in a wool cap and dark coat gazing at the open sea, with lanterns swaying gently from the rigging
  \item A lone sailor in a wool cap and dark coat standing on the deck smiling as the ship approaches a massive arch of stone rising
    from the water
  \item A lone sailor in a wool cap and dark coat resting beside the ship's steering wheel, moonlight shimmering on the sea
  \item A lone sailor in a wool cap and dark coat guiding the ship through towering jade waves as bioluminescent whales breach alongside
  \item A lone sailor in a wool cap and dark coat playing a low whistle tune on the bow beneath a sky ablaze with meteoric shards
\end{enumerate}

This structured prompt ensures that GPT-4o generates visually grounded, narratively coherent storyboards--each with a central character, evolving setting, and cinematic composition. The Rich Storyboard Benchmark allows us to systematically test a model’s ability to maintain identity while navigating scene transitions and visual storytelling demands.

\subsection{Scene Diversity Metric}
\label{sec:appendix-scenediversity}

To quantify layout variation in a storyboard, we introduce the \textbf{Scene Diversity} metric. It captures how dynamically a subject is presented across panels--accounting for changes in framing, position, and pose.

Given a storyboard of $n$ panels and a text description identifying the subject, we locate the subject in each image using Grounding DINO~\cite{Oquab2023DINOv2LR}. For each panel, we extract a bounding box around the subject and normalize it by the image dimensions. We then compute the per-coordinate standard deviation of these normalized bounding boxes across the $n$ panels, and average the result. This yields the \textit{bounding box std score}, which reflects variation in subject placement and scale.

For stories with human characters, we additionally compute 17 pose keypoints per panel using ViTPose. We calculate the per-keypoint variance across all panels and average them to obtain a \textit{pose variance score}, representing variation in articulation and body posture.

Each score is then min-max normalized across all stories in the benchmark. We denote the normalized bounding box and pose scores as $s_{\text{bbox}}$ and $s_{\text{pose}}$ respectively. The final scene diversity score is computed as:
\[
\text{Scene Diversity} = \begin{cases}
    \frac{1}{2}(s_{\text{bbox}} + s_{\text{pose}}), & \text{if human} \\
    s_{\text{bbox}}, & \text{otherwise}
\end{cases}
\]

This metric enables us to evaluate a model’s ability to vary subject framing and presentation across narrative beats, a core requirement for expressive visual storytelling.

\subsection{Implementation Details}
\label{sec:appendix-implementation}

We implement our method using the open-source Flux diffusion model~\cite{flux}. All experiments are conducted on the pre-trained Flux.1-dev model without additional training or finetuning.

\paragraph{Reference Panel Selection.}
As described in the main paper, our pipeline requires a single \textit{reference panel} to serve as the source of character identity features. In practice, we reuse one of the storyboard panels that already includes all the main characters--typically the first or second scene in the sequence. We found that this approach produces results as good as or better than using a separately rendered reference panel, while also saving inference time and VRAM.

\paragraph{RAVM Details.}
We apply Reciprocal Attention Value Mixing (RAVM) at inference time using the following configuration:
\begin{itemize}
  \item We run inference with classifier-free guidance of 3.5 and 28 denoising steps.
  \item RAVM is applied to all 38 dual-stream transformer blocks in Flux’s denoising network.
  \item We use a mixing parameter $\lambda = 0.5$ to control the blend between source and target token values.
  \item The reciprocal attention maps are smoothed using exponential decay with a momentum of $0.8$.
\end{itemize}

Our full pipeline is training-free, fast to run, and requires no architectural modification to the underlying diffusion transformer. All interventions are performed at the attention value level during sampling.

\subsection{DS-500 Evaluation}
\label{sec:appendix-ds-500}
To assess generalization beyond our benchmark, we also evaluate on DS-500~\cite{he2024dreamstory}, a storyboard dataset with shorter prompts and minimal scene evolution. While not designed to test layout or narrative expressivity, DS-500 serves as a useful baseline for identity coherence and basic prompt alignment.

\begin{table}[h]
\centering
\small
\begin{tabular}{lcc}
\toprule
\textbf{Method} & \textbf{CLIP-T (↑)} & \textbf{DreamSim (↑)} \\
\midrule
DreamStory~\cite{he2025anystory} & \textbf{0.3779} & 0.6714 \\
\textbf{Story2Board (ours)} & 0.3723 & \textbf{0.7018} \\
\bottomrule
\end{tabular}
\vspace{4pt}
\caption{DS-500 evaluation results. DreamStory's scores are reported directly from their paper~\cite{he2025anystory}. Our method achieves competitive CLIP-T alignment while outperforming DreamStory in identity consistency (DreamSim).}
\label{tab:ds500}
\end{table}

As shown in Table~\ref{tab:ds500}, our method achieves comparable prompt alignment (CLIP-T)~\cite{OpenCLIP} and higher identity consistency (DreamSim~\cite{Fu2023DreamSimLN}) relative to DreamStory~\cite{he2024dreamstory}. This supports the broader applicability of our approach across datasets with varying narrative complexity.

\subsection{User Study Details}
\label{sec:appendix-userstudy}

To evaluate the perceived quality of generated storyboards, we conducted a large-scale user study on Amazon Mechanical Turk (MTurk)~\cite{amazonmturk} using all 100 stories from our Rich Storyboard Benchmark. We used the first four storyboard panels from each story as rendered by our method and one competing baseline. The resulting pairs were shown to participants side-by-side.

\paragraph{Study Design.}
We ran five separate studies, each targeting a specific evaluation criterion:
\begin{enumerate}
  \item \textbf{Overall Preference}
  \item \textbf{Prompt Alignment}
  \item \textbf{Character Consistency}
  \item \textbf{Background Richness}
  \item \textbf{Scene Diversity}
\end{enumerate}

To ensure consistent comparison coverage across baselines, we divided the 100 stories into five disjoint groups of 20 stories. Each group was assigned to a different baseline, resulting in one set of storyboards per competitor for evaluation against Story2Board.

\paragraph{Participant Selection.}
To ensure high-quality responses, we restricted participation to workers located in English-speaking countries--specifically the United States, United Kingdom, Canada, and Australia. We further filtered for workers with a lifetime task approval rate above 98\%, prioritizing reliable and experienced annotators.

\paragraph{Interface and Task.}
For each evaluation criterion, participants were shown a sequence of trials. Each trial displayed two storyboards (A and B) generated from the same story prompt--one from Story2Board and one from a competing model. Participants were asked to select the storyboard that best satisfied the target criterion. Model names and ordering were not shown.

Each trial presented all four storyboard panels per model, along with their associated captions. Image layouts were standardized and left–right positioning was randomized. Each trial was rated by 3 unique workers.

\paragraph{Instructions to Participants.}
Participants were given the following instructions at the start of each task. The only variation across studies was the criterion description, shown in bold.

\begin{figure}[t]
  \centering
  \includegraphics[width=\textwidth]{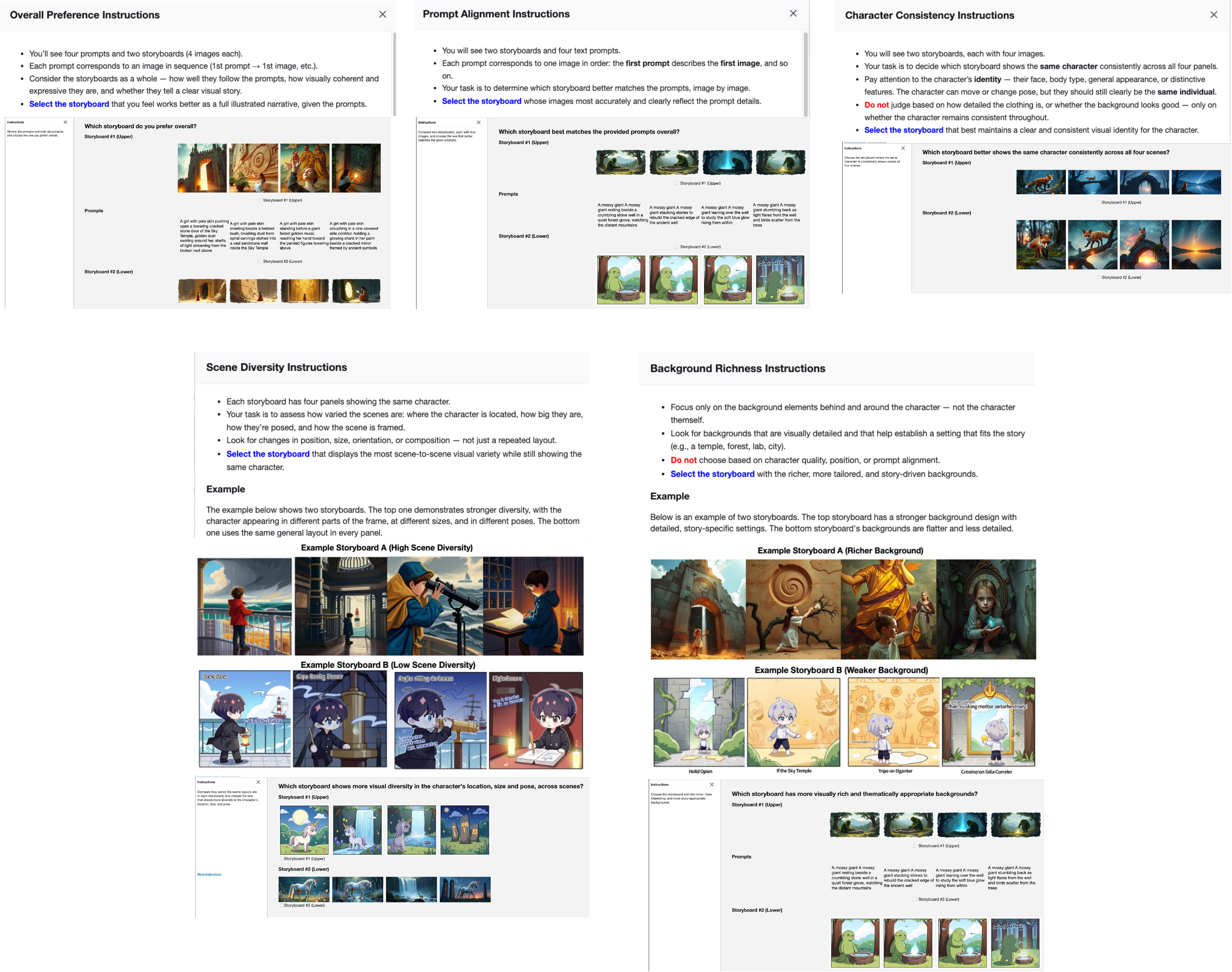}
  \caption{User Study Instructions. We provide the complete instructions for the user study we conducted using Amazon Mechanical Turk (AMT) to compare our method with each baseline.}
  \label{fig:user_study_examples}
\end{figure}

For other criteria, the bolded instruction was replaced accordingly. For instance, for \textit{Prompt Alignment}, participants were asked to choose the version that more accurately matched the text descriptions; for \textit{Scene Diversity}, they were asked to consider how much variety was present across the panels in terms of framing, layout, and setting. Screenshots are presented in Figure~\ref{fig:user_study_examples}.

For criteria that required more subjective interpretation--such as \textit{Scene Diversity} and \textit{Background Richness}--participants were also shown example pairs of good and poor storyboards illustrating the concept, drawn from baseline methods and distinct stories not used in the evaluation.

\paragraph{Result Aggregation.}
Participant responses were aggregated across all trials per criterion to compute win rates. These results are summarized in Figure~\ref{fig:user-study} in the main paper. Our method received the highest overall preference scores, winning the majority of pairwise comparisons in the “Overall Preference” category. This indicates that when participants considered the storyboards as a whole, they consistently favored our approach over all competitors.

At the same time, the results highlight specific trade-offs across individual evaluation criteria. OminiControl achieved stronger scores in prompt alignment, background richness, and scene diversity, likely benefiting from its encoder-based layout conditioning. Meanwhile, IC-LoRA (Storyboards) and StoryDiffusion were slightly favored for character consistency, reflecting their targeted emphasis on identity preservation. In contrast, our method’s use of soft, token-level guidance enables greater flexibility in layout and framing--traits that may account for its overall appeal despite falling behind in some focused categories.

\FloatBarrier
\onecolumn

\subsection{Full Story Texts}
\label{sec:appendix-full-story-texts}
See Table~\ref{table:full-story-texts} for examples of complete stories.
\vspace{1em}

\noindent
\captionof{table}{Full Story Texts}\label{table:full-story-texts}
\renewcommand{\arraystretch}{1.2}
\begin{tabularx}{\textwidth}{@{}p{0.25\textwidth} p{0.72\textwidth}@{}}
\toprule
\textbf{Storyboard ID} & \textbf{Full Story Text} \\
\midrule

Rami the Desert Nomad &
Rami, the desert nomad, held his glowing lantern high beside his patient camel, the twilight painting the dunes in shades of blood and gold. The endless sands stretched away into eternity, broken only by whispers of ancient paths.

As he pressed forward, Rami passed beneath a ruined sandstone arch, its surface etched by time. Crimson petals blew past in sudden gusts, swirling around him like lost memories, as if the gate itself were exhaling the past.

That night, silhouetted against a crescent moon atop a dune ridge, Rami paused. The wind tugged at his cloak while the cold stars wheeled above, casting long shadows across the rippling sand.

Seeking shelter, he settled beneath a jagged stone outcrop, kneeling on worn rock. By the glow of his lantern, he unrolled a crumpled map, squinting at the faded lines and markings, unsure of what was memory and what was myth.

At dawn, a caravan of dune-moth herders emerged from the haze of a violet dust storm. Rami approached cautiously, negotiating passage through their shifting territory. Strange banners fluttered from their saddles; their moths blinked with luminous eyes.

By dusk, he reached a tower of bones--an ancient, impossible spire that pierced the desert sky. Rami climbed its spiraling ramp, each step echoing with forgotten oaths, until he stood at its apex.

There, atop a glassy dune, Rami raised his lantern one final time. Its golden glow danced against the wind as twin moons rose behind him. In the distance, tiny signals flickered in reply--other wanderers answering his call across the sands. \\

Blackpaw in the Celestial Forest &
Blackpaw, a shimmering fox of the ancient celestial forest, stepped lightly
onto a mossy stone path, the twilight trees arching high above him like a cathedral
ceiling.

With a flick of his glowing tail, he bounded across a fallen tree stretched precariously
over a mist-shrouded ravine that gleamed faintly with constellations reflected
in the fog below.

Perched atop a broken archway of ancient stone, vines and silver moss hanging
around him, Blackpaw gazed out over the glowing forest as twilight deepened.
From the edge of a luminous lake mirroring the heavens perfectly, he watched a
meteor shower ignite the sky, each fiery streak mirrored twice over.

Curling beside a pulsing crystal monolith, he dreamed in the ancient heartbeat
of the forest. By morning, the grove was silent but for whispering silver leaves
shedding light into the wind, and the fading trace of the fox's gleaming trail. \\

The Last Astronomer &
They called her Dr. Elira Voss, though no one had used her title in years. She was the last custodian of the Skyreach Observatory, a rusting dome perched on the cliffs where stars once spoke to science. Beneath its cracked shutters and wind-scoured walls, Elira still watched the sky--not for data, but for memory.

A single tear traced her cheek as a meteor shower flared across the heavens, scattering silver sparks over the dark sea. She stood silently beside a weathered telescope, its brass fittings dulled by time, and turned back to her hand-drawn sky chart. The map was crowded with inked constellations, margins lined with notes and dates only she understood.

She moved carefully, peering through the eyepiece and adjusting the scope until a distant galaxy came into view. Her fingers trembled, not from age, but from the echo of another life. On the desk nearby sat a faded photograph of a man in an astronaut’s suit--his smile still intact, his absence louder than ever.

Later that night, she spotted it: a new star, impossibly bright. Her breath caught. She smiled--not wide, not triumphant, but soft, as if welcoming an old friend. She cranked the observatory’s rusted gears, pulling open the cracked dome just in time to follow a teal comet slicing across the sky, its fire washing over ancient machinery like a blessing.

And when her hands could do no more, she stepped onto the rooftop and lit a paper lantern. As it rose, its glow joined the glinting trail of satellite beacons. A message. A memory. A promise that she was still watching, still waiting. Still listening. \\

\bottomrule
\end{tabularx}

\end{document}